\RequirePackage{fix-cm}
\documentclass[twocolumn]{svjour3}          % twocolumn
\smartqed  % flush right qed marks, e.g. at end of proof
\usepackage{graphicx}
\usepackage{makecell}
\usepackage{amsmath,amssymb,amsfonts}
\usepackage{algorithmic}
\usepackage{graphicx}
\usepackage{textcomp}
\usepackage{xcolor}
\usepackage[title]{appendix}
\usepackage{array,multirow}
\usepackage[font=small]{caption}
\usepackage{balance}  % for  \balance command ON LAST PAGE  (only there!)

\usepackage{color, colortbl}
\definecolor{Gray}{gray}{0.8}
\definecolor{lGray}{gray}{0.9}
\usepackage{multicol}

\usepackage{enumitem}

\usepackage[linesnumbered,algoruled,boxed,lined,noend]{algorithm2e}

\SetCommentSty{mycommfont}

\newcommand{\commentRedXMARK}[1]{{\color[rgb]{1,0,0}#1}}

\usepackage{comment}
\usepackage{url}
\usepackage{tikz}
\newcommand{\commentRed}[1]{{\color[rgb]{0,0,0}~#1}}
\newcommand{\greencheck}{}%
\DeclareRobustCommand{\greencheck}{%
\textbf{
  \tikz\fill[scale=0.4, color=green]
  (0,.35) -- (.25,0) -- (1,.7) -- (.25,.15) -- cycle;%
}}
\usepackage{pifont}
\newcommand{\xmark}{\text{\ding{55}}}
\newcommand{\redmark}{\text{\commentRedXMARK{\xmark}}}%
\begin{document}

\title{VUS: Effective and Efficient Accuracy Measures for Time-Series Anomaly Detection%\thanks{Grants or other notes
}


\author{Paul Boniol \and
        Ashwin K. Krishna \and
        Marine Bruel \and
        Qinghua Liu \and 
        Mingyi Huang \and 
        Themis Palpanas \and
        Ruey S. Tsay \and
        Aaron Elmore \and
        Michael J. Franklin \and
        John Paparrizos \and%etc.
}


\institute{Paul Boniol \at
              45 rue d'Ulm, 75005, Paris \\
              \email{boniol.paul@gmail.com}
}

\date{Received: date / Accepted: date}

\maketitle

\begin{abstract}
Anomaly detection (AD) is a fundamental task for time-series analytics with important implications for the downstream performance of many applications. In contrast to other domains where AD mainly focuses on point-based anomalies (i.e., outliers in standalone observations), AD for time series is also concerned with range-based anomalies (i.e., outliers spanning multiple observations). Nevertheless, it is common to use traditional point-based information retrieval measures, such as Precision, Recall, and F-score, to assess the quality of methods by thresholding the anomaly score to mark each point as an anomaly or not. However, mapping discrete labels into continuous data introduces unavoidable shortcomings, complicating the evaluation of range-based anomalies. Notably, the choice of evaluation measure may significantly bias the experimental outcome. Despite over six decades of attention, there has never been a large-scale systematic quantitative and qualitative analysis of time-series AD evaluation measures. This paper extensively evaluates quality measures for time-series AD to assess their robustness under noise, misalignments, and different anomaly cardinality ratios. Our results indicate that measures producing quality values independently of a threshold (i.e., AUC-ROC and AUC-PR) are more suitable for time-series AD. Motivated by this observation, we first extend the AUC-based measures to account for range-based anomalies. Then, we introduce a new family of parameter-free and threshold-independent measures, Volume Under the Surface (VUS), to evaluate methods while varying parameters. We also introduce two optimized implementations for VUS that reduce significantly the execution time of the initial implementation. Our findings demonstrate that our four measures are significantly more robust in assessing the quality of time-series AD methods.
%\keywords{Time Series \and Anomaly Detection \and Evaluation Measures}
\end{abstract}

\begin{figure}
 \centering
 \includegraphics[width=\linewidth]{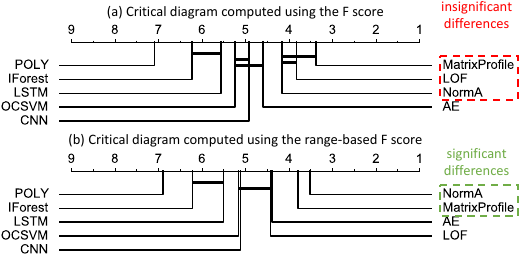}
 \caption{Critical difference diagram computed with the Friedman test followed by a post-hoc Wilcoxon test (with $\alpha=0.1$) for the (a) F-score and (b) range-based F-score over $250$ time series in KDD21 \cite{kdd}. Bold lines indicate insignificant differences of connected methods.}
 \label{fig:introf}
\end{figure}

\section{Introduction}

Massive collections of time-varying measurements, commonly referred to as \textit{time series}, have become a reality in virtually every scientific and industrial domain ~\cite{DBLP:journals/dagstuhl-reports/BagnallCPZ19,Palpanas2019,paparrizos_k-shape_2016,paparrizos2019grail,paparrizos2020debunking,dziedzic2019band,bariya2021k,paparrizos22fast,paparrizos2018fast}. Notably, there is an increasingly pressing need for developing techniques for efficient and effective analysis of zettabytes of time series produced by millions of Internet-of-Things (IoT) devices \cite{iotstats,hung2017leading,paparrizos2021vergedb,jiang2020pids,jiang2021good,liu2021decomposed}. 
IoT deployments empower diverse data science applications in environmental sciences, astrophysics, neuroscience, and engineering, among others~\cite{Palpanas2015,fulfillingtheneed}, and have revolutionized many industries, including automobile, healthcare, manufacturing, and utilities~\cite{ng2017internet}. 
However, rare events, or imperfections and inherent complexities in the data generation and measurement pipelines, often introduce abnormalities that appear as \textit{anomalies} in time-series databases, impacting the effectiveness of downstream tasks and analytics.

Consequently, \textit{anomaly detection} (AD) becomes a fundamental problem with broad applications sharing the same goal~\cite{statisticaloutliers,DBLP:conf/vldb/SubramaniamPPKG06,DBLP:conf/icdm/YehZUBDDSMK16}: analyzing time series to identify observations that do not conform to some notion of expected behavior based on previously observed data. During the past decades, a multitude of AD methods have been proposed and compared~\cite{yeh_time_2018,DBLP:journals/datamine/LinardiZPK20,DBLP:conf/icde/BoniolLRP20a,DBLP:conf/edbt/Gao0B20,boniol_unsupervised_2021,boniol2021sandaction,Series2GraphPaper,DBLP:journals/pvldb/BoniolPPF21,DBLP:conf/icdm/YehZUBDDSMK16,Liu:2008:IF:1510528.1511387,breunig_lof_2000,DBLP:journals/csur/Blazquez-Garcia21,DBLP:journals/pvldb/PaparrizosKBTPF22,theseus}. Different from other domains that principally focus on \textit{point-based} anomalies (i.e., outliers in standalone observations), AD for time series is also concerned with \textit{range-based} anomalies (i.e., outliers spanning multiple observations). 
Unfortunately, it has become common practice to use traditional point-based information retrieval (IR) accuracy measures, such as Precision, Recall, and F-score, to quantify the effectiveness of different anomaly detectors.

In addition, the previously mentioned IR evaluation measures suffer from a significant limitation: a threshold is necessary over the anomaly score produced by AD methods to mark each time-series point as an anomaly or not. The most common approach to set a threshold value is to use the average score plus three times the standard deviation of the anomaly score. However, this popular choice might not suit every AD method, use case, and domain, leading to significant variations in the quality values of the evaluation measures. Therefore, these IR measures are difficult to trust and complicate evaluating different AD methods on heterogeneous benchmarks. To eliminate the need to set a threshold, another standard measure for binary classification is used: the receiver operator characteristic (ROC) curve and the Area Under the Curve (AUC), which is the area below the ROC curve (AUC-ROC). The ROC curve is generated by plotting the true positive rate (TPR) against the false positive rate (FPR) at various threshold settings (instead of only one threshold used in Precision, Recall, and F-score measures). Another variant, the Precision-Recall (PR) curve, represents the relation between Precision and Recall, and the Area under the PR curve (AUC-PR) is the area below PR \cite{10.1145/1143844.1143874}.

Unfortunately, all previous measures, Precision, Recall, F-Score, AUC-ROC, and AUC-PR, are ideal for point-based anomalies but cannot adequately evaluate ubiquitous range-based contextual and collective anomalies \cite{blazquez2021review}. Remarkably, the mapping of discrete labels into continuous data introduces unavoidable shortcomings (e.g., difficulty in marking precisely the range of the anomalies and handling misalignments between the human labels and the anomaly range produced by thresholding the anomaly score). To address these shortcomings, a range-based definition of Precision and Recall has been proposed by extending the traditional definitions \cite{tatbul_precision_2018}. Range-based Precision, Recall, and F-Score consider several factors: (i) whether a subsequence is detected or not; (ii) how many points in the subsequence are detected; (iii) which part of the subsequence is detected; and (iv) how many fragmented regions correspond to one real subsequence outlier. This definition is detailed and comprehensive; however, several parameters require tuning and, importantly, a threshold over the anomaly score is still required.

A recent study~\cite{DBLP:journals/corr/abs-2303-01272} listed AD evaluation measures for time series, describing their advantages and shortcomings measured on synthetic time series. However, there has never been (to the best of our knowledge) a large-scale systematic quantitative and qualitative analysis of time-series AD evaluation measures on real time series. Notably, the choice of evaluation measure may significantly bias the experimental outcome. To understand the implications of choosing an appropriate measure, Figure \ref{fig:introf} depicts the critical diagrams of the F-score and range-based F-score computed with the Friedman test followed by a Wilcoxon test~\cite{10.2307/3001968} over several AD methods (see Section \ref{sec:exp} for details) across the $250$ time series of the KDD21 dataset~\cite{kdd}. Figure~\ref{fig:introf} demonstrates that not only the ranking is changing, but also some methods shift from insignificantly to significantly different from one measure to the other.

In this paper, we extensively evaluate quality measures for time-series AD to assess their robustness under noise, misalignments, and different anomaly cardinality ratios. Specifically, our study includes $9$ previously proposed quality measures, computed over the anomaly scores of $10$ AD methods across $10$ diverse datasets that contain $900$ time series with marked anomalies. Our analysis assesses the robustness of quality measures both qualitatively and quantitatively by studying the influence of threshold, lag, noise, and normal-abnormal anomaly ratio to identify robust measures that better separate accurate from inaccurate methods. 

Our results indicate that measures producing quality values independently of a threshold (i.e., AUC-ROC and AUC-PR) are more suitable for time-series AD. This is surprising considering that we include the range-based Precision, Recall, and F-score measures, which highlights the strong influence the thresholding of anomaly scores has in assessing the quality of methods. 

Motivated by this observation and to address the limitations of existing measures, we propose \textit{four} new accuracy evaluation measures. We first present Range-AUC-ROC and Range-AUC-PR, threshold-independent (for the anomaly score) evaluation measures that use a continuous buffer region in the labels to increase the robustness to potential misalignments with the human labels. Then, we propose the Volume Under the Surface (VUS) family of measures that extend the traditional AUC measures to consider all buffer sizes (in addition to all thresholds). Therefore, VUS-ROC and VUS-PR are parameter-free, threshold-independent, and robust to lags, noise, and anomaly cardinality ratios. 
Our analysis demonstrates that VUS-ROC and VUS-PR are the most reliable accuracy quality measures for both point-based and range-based anomaly evaluation. Table~\ref{methodTable} summarizes the accuracy evaluation measures analyzed in this paper based on their independence to four critical characteristics.

In addition to the accuracy evaluation, we perform an extensive execution time evaluation. 
VUS requires computing accuracy measures (i.e., ROC, Precision, and Recall) for different values of buffer sizes.
As this buffer size changes the labels of the time series, the naive implementation of VUS computes accuracy measures over the entire labels as many times as the number of buffer sizes we consider. However, the buffer size affects only small sections of the labels, leaving the vast majority unchanged. Therefore, we introduce two optimized versions of the VUS computation algorithm that compute accuracy measures over the sections affected by the buffer length. We demonstrate theoretically and empirically the execution time improvement of the optimized implementations over the naive implementation of VUS while remaining exact (i.e., providing the same values as the naive implementation). Overall, our optimized implementations is up to 10 times faster than the naive implementation for large time series, and render the VUS measures easier to use in practice.

\begin{table}[tp]
\caption{Analysis of quality measures based on: (i) independence to the number of anomalies; (ii) independence to the threshold; (iii) adaptation to continuous sequences; and (iv) independence to setting parameters.}
\label{methodTable}
\centering
\scalebox{0.70}{
\begin{tabular}{|c||c|c|c|c|}
\hline
{\bf Acc. Measure} & \# of anom. & Score Thres. & Sequence-adapted & Param-free \\
\hline
Precision@k & \redmark & \greencheck & \redmark & \redmark \\
Precision & \greencheck & \redmark & \redmark & \redmark \\
Recall & \greencheck & \redmark & \redmark & \redmark \\
F-Score & \greencheck & \redmark & \redmark & \redmark\\
Rprecision & \greencheck & \redmark & \greencheck & \redmark \\
Rrecall & \greencheck & \redmark & \greencheck & \redmark \\
RF-Score & \greencheck & \redmark & \greencheck & \redmark \\
AUC-PR & \greencheck & \greencheck & \redmark &\greencheck \\
AUC-ROC & \greencheck & \greencheck & \redmark & \greencheck \\
\hline
\rowcolor{Gray}
\multicolumn{5}{|c|}{{\bf \emph{Proposed measures}}}\\
\hline
R-AUC-PR & \greencheck & \greencheck & \greencheck & \redmark \\
R-AUC-ROC & \greencheck & \greencheck & \greencheck & \redmark \\
VUS-PR & \greencheck & \greencheck & \greencheck & \greencheck \\
VUS-ROC & \greencheck & \greencheck & \greencheck & \greencheck \\
\hline
\end{tabular}
} % scalebox
\end{table}

Interestingly, even though outside of the scope of this paper, the flexibility of VUS measures in evaluating methods while varying parameters of choice may have implications beyond time-series AD. Specifically, VUS measures are applicable across binary classification tasks for evaluating methods with a single quality value while considering different parameter choices (e.g., learning rates, batch sizes, and other critical varying parameters). 

\noindent{\bf (Sec.~\ref{sec:background})} We start with a detailed discussion of the relevant background and related work. Then, we present our contributions\footnote{A preliminary version has appeared elsewhere~\cite{DBLP:journals/pvldb/PaparrizosBPTEF22}.}:

\noindent{\bf (Sec.~\ref{sec:problem})} We discuss the limitations of existing evaluation measures, resulting in a formal definition of the necessary principles of time-series AD quality measures.

\noindent{\bf (Sec.~\ref{sec:range-auc})} We present R-AUC (ROC and PR) that rely on a new label transformation for a more robust and reliable score for contextual and collective anomalies.

\noindent{\bf (Sec.~\ref{sec:vus})} We introduce VUS (ROC and PR), parameter-free measures that formally extend AUC-based measures to consider more varying parameters.

\noindent{\bf (Sec.~\ref{sec:fasterimpl})} We introduce $VUS_{opt}$ and $VUS^{mem}_{opt}$, two optimized versions for the computation of both VUS-ROC and VUS-PR, with significantly better time complexity properties. These two optimized versions prune the sections of the time series in which the anomaly score does not change regardless of the threshold and the buffer length. The $VUS^{mem}_{opt}$ algorithm further improves time-complexity by using more memory.

\noindent{\bf (Sec.~\ref{exp:qual} and~\ref{exp:quant})} We extensively evaluate, both qualitatively and quantitatively, $13$ quality measures ($9$ previously proposed and our $4$ new measures) across $10$ AD methods over $10$ diverse datasets containing $900$ time series with marked anomalies.

\noindent{\bf (Sec.~\ref{exp:separability})} We analyze the separability of the measures by comparing pairs of accurate and inaccurate methods.

\noindent{\bf (Sec.~\ref{sec:entropy})} We assess the consistency of the measures by evaluating changes in methods' ranks across measures.

\noindent{\bf (Sec.~\ref{sec:exectime})} We evaluate the scalability of the VUS-based measures on different time series characteristics, and we measure the speed-up of $VUS_{opt}$ and $VUS^{mem}_{opt}$ compared to the naive implementation of VUS.

\noindent{\bf (Sec.~\ref{sec:conclusions})} Finally, we conclude with the implications of our work and discuss future research directions.

\section{Background and Related Work}
\label{sec:background}

We first introduce formal notations useful for the rest of the paper (Section \ref{sec:notation}). Then, we review in detail previously proposed evaluation measures for time-series AD methods (Section \ref{acc_measure}).

\subsection{Time-Series and Anomaly Score Notations}
\label{sec:notation}
We review notations for the time series and anomaly score sequence.
\newline \textbf{Time Series: } A time series $T \in \mathbb{R}^n $ is a sequence of real-valued numbers $T_i\in\mathbb{R}$ $[T_1,T_2,...,T_n]$, where $n=|T|$ is the length of $T$, and $T_i$ is the $i^{th}$ point of $T$. We are typically interested in local regions of the time series, known as subsequences. A subsequence $T_{i,\ell} \in \mathbb{R}^\ell$ of a time series $T$ is a continuous subset of the values of $T$ of length $\ell$ starting at position $i$. Formally, $T_{i,\ell} = [T_i, T_{i+1},...,T_{i+\ell-1}]$.	
\newline \textbf{Anomaly Score Sequence: } For a time series $T \in \mathbb{R}^n $, an AD method $A$ returns an anomaly score sequence $S_T$. For point-based approaches (i.e., methods that return a score for each point of $T$), we have $S_T \in \mathbb{R}^n$. For range-based approaches (i.e., methods that return a score for each subsequence of a given length $\ell$), we have $S_T \in \mathbb{R}^{n-\ell}$. Overall, for range-based (or subsequence-based) approaches, we define $S_T = [{S_T}_1,{S_T}_2,...,{S_T}_{n-\ell}]$ with ${S_T}_i \in [0,1]$.

\subsection{Accuracy Evaluation Measures for AD}
\label{acc_measure}

We present previously proposed quality measures for evaluating the accuracy of an AD method, given its anomaly score. We first discuss threshold-based and then threshold-independent measures.

\subsubsection{Threshold-based AD Evaluation Measures} \hfill\\
The anomaly score $S_T$ produced by an AD method $A$ highlights the parts of the time series $T$ considered as abnormal. The highest values in the anomaly score correspond to the most abnormal points. Threshold-based measures require setting a threshold to mark each point as an anomaly or not. Usually, this threshold is set to $\mu(S_T) + \alpha*\sigma(S_T)$, with $\alpha$ set to 3~\cite{statisticaloutliers}, where $\mu(S_T)$ is the mean and $\sigma(S_T)$ is the standard deviation $S_T$. Given a threshold $Thres$, we compute the $pred \in \{0,1\}^n$ as follows:

\begin{equation}
\begin{split}
&\forall i \in [1,|S_T|], pred_i = \left.
\begin{cases}
0,& \text{if: } {S_T}_i < Thres \\
1,& \text{if: } {S_T}_i \geq Thres 
\end{cases}
\right.
\end{split}
\end{equation}

Threshold-based measures compare $pred$ to $label \in \{0,1\}^n$, which indicates the true (human provided) labeled anomalies. Given the identity vector $I=[1,1,...,1]$, the points detected as anomalies or not fall into the following four categories:
\begin{itemize}[noitemsep,topsep=0pt,parsep=0pt,partopsep=0pt,leftmargin=0.5cm]
	\item {\bf True Positive (TP)}: Number of points that have been correctly identified as anomalies. Formally: $TP = label^\top \cdot pred$.
	\item {\bf True Negative (TN)}: Number of points that have been correctly identified as normal. Formally: $TN = (I-label)^\top \cdot (I-pred)$.
	\item {\bf False Positive (FP)}: Number of points that have been wrongly identified as anomalies. Formally: $FP = (I-label)^\top \cdot pred$.
	\item {\bf False Negative (FN)}: Number of points that have been wrongly identified as normal. Formally: $FN = label^\top \cdot (I-pred)$.
\end{itemize}
Given these categories, several quality measures have been proposed to assess the accuracy of AD methods.
\newline \textbf{Precision: } We define Precision (or positive predictive value) as the number of correctly identified anomalies over the total number of points detected as anomalies by the method:

\begin{equation}
Precision = \frac{TP}{TP+FP}
\end{equation}
\newline \textbf{Recall: } We define Recall (or True Positive Rate (TPR), $tpr$) as the number of correctly identified anomalies over all anomalies:

\begin{equation}
Recall = \frac{TP}{TP+FN}
\end{equation}

\noindent \textbf{False Positive Rate (FPR): } A supplemental measure to the Recall is the FPR, $fpr$, defined as the number of points wrongly identified as anomalies over the total number of normal points:

\begin{equation}
fpr = \frac{FP}{FP+TN}
\end{equation}
\newline \textbf{F-Score: } Precision and Recall evaluate two different aspects of the AD quality. A measure that combines these two aspects is the harmonic mean $F_{\beta}$, with non-negative real values for $\beta$:
\begin{equation}
F_{\beta} = \frac{(1+\beta^2)*Precision*Recall}{\beta^2*Precision+Recall}
\end{equation}
\noindent Usually, $\beta$ is set to 1, balancing the importance between Precision and Recall. In this paper, $F_1$ is referred to as F or F-score.
\newline \textbf{Precision@k: } All previous measures require an anomaly score threshold to be computed. An alternative approach is to measure the Precision using a subset of anomalies corresponding to the $k$ highest value in the anomaly score $S_T$. This is equivalent to setting the threshold such that only the $k$ highest values are retrieved. 

To address the shortcomings of the point-based measures, a range-based definition was proposed, extending the traditional Precision and Recall \cite{tatbul_precision_2018}. This definition considers several factors: (i) whether a subsequence is detected or not (ExistenceReward or ER); (ii) how many points in the subsequence are detected (OverlapReward or OR); (iii) which part of the subsequence is detected (position-dependent weight function); and (iv) how many fragmented regions correspond to one real subsequence outlier (CardinalityFactor or CF). Formally, we define $R=\{R_1,...R_{N_r}\}$ as the set of anomaly ranges, with $R_k=\{pos_i,pos_{i+1}, ..., pos_{i+j}\}$ and $\forall pos \in R_k, label_{pos} = 1$, and $P=\{P_1,...P_{N_p}\}$ as the set of predicted anomaly ranges, with $P_k=\{pos_i,pos_{i+1}, ..., pos_{i+j}\}$ and $\forall pos \in R_k, pred_{pos} = 1$. Then, we define ER, OR, and CF as follows:

%\vspace{-0.3cm}
\begin{equation}
\footnotesize{
\begin{split}
&ER(R_i,P) = \left.
\begin{cases}
1, &\text{if } \sum_{j=1}^{N_p} |R_i \cap P_j| \geq 1\\
0, &\text{otherwise}
\end{cases}
\right. \\
&CF(R_i,P) = \left.
\begin{cases}
1, &\text{if } \exists P_i \in P, |R_i \cap P_i| \geq 1\\
\gamma(R_i,P), &\text{otherwise}
\end{cases}
\right. \\
&OR(R_i,P) = CF(R_i,P)*\sum_{j=1}^{N_p} \omega(R_i,R_i \cap P_j, \delta)
\end{split}
}
%\vspace{-0.1cm}
\end{equation}

\noindent The $\gamma(),\omega()$, and $\delta()$ are tunable functions that capture the cardinality, size, and position of the overlap respectively.
The default parameters are set to $\gamma()=1,\delta()=1$ and $\omega()$ to the overlap ratio covered by the predicted anomaly range~\cite{tatbul_precision_2018}.
\newline \textbf{Rprecision and Rrecall~\cite{tatbul_precision_2018}: } Based on the above, we define:

%\vspace{-0.2cm}
\begin{equation}
\footnotesize{
\begin{split}
Rprecision(R,P) &= \frac{\sum_{i=1}^{N_p} Rprecision_s(R,P_i)}{N_p}\\
Rprecision_s(R,P_i) &= CF(P_i,R)*\sum_{j=1}^{N_r} \omega(P_i,P_i \cap R_j, \delta)
\end{split}
}
%\vspace{-0.2cm}
\end{equation}

%%\vspace{-0.2cm}
\begin{equation}
\footnotesize{
\begin{split}
Rrecall(R,P) &= \frac{\sum_{i=1}^{N_r} Rrecall_s(R_i,P)}{N_r} \\
Rrecall_s(R_i,P) &= \alpha*ER(R_i,P) + (1-\alpha)*OR(R_i,P)
\end{split}
}
%%\vspace{-0.1cm}
\end{equation}
\noindent The parameter $\alpha$ is user defined. The default value is $\alpha=0$.
\newline \textbf{Range F-score (RF)~\cite{tatbul_precision_2018}: } As described previously, the F-score combines Precision and Recall. Similarly, we define $RF_{\beta}$, for $\beta>0$ as follows:

%\vspace{-0.3cm}
\begin{equation}
RF_{\beta} = \frac{(1+\beta^2)*Rprecision*Rrecall}{\beta^2*Rprecision+Rrecall}
%\vspace{-0.1cm}
\end{equation}

\noindent As before, $\beta$ is set to 1. In this paper, $RF_1$ is referred to as RF-score.

\subsubsection{Threshold-independent AD Evaluation Measures} \hfill\\
Until now, we introduced accuracy measures requiring to threshold the produced anomaly score of AD methods. However, the accuracy values vary significantly when the threshold changes. To evaluate a method holistically using its corresponding anomaly score, two measures from the AUC family of measures are used.
\newline \textbf{AUC-ROC~\cite{FAWCETT2006861}: } The Area Under the Receiver Operating Characteristics curve (AUC-ROC) is defined as the area under the curve corresponding to TPR on the y-axis and FPR on the x-axis when we vary the anomaly score threshold. The area under the curve is computed using the trapezoidal rule. For that purpose, we define $Th$ as an ordered set of thresholds between 0 and 1. Formally, we have $Th=[Th_0,Th_1,...Th_N]$ with $0=Th_0<Th_1<...<Th_N=1$. Therefore, $AUC\text{-}ROC$ is defined as follows:
%\vspace{-0.2cm}
\begin{equation}
\begin{split}
&AUC\text{-}ROC = \frac{1}{2}\sum_{k=1}^{N} \Delta^{k}_{TPR}*\Delta^{k}_{FPR}\\
&\text{with: } \left.
\begin{cases}
\Delta^{k}_{FPR} &= FPR(Th_{k})-FPR(Th_{k-1})\\
\Delta^{k}_{TPR} &= TPR(Th_{k-1})+TPR(Th_{k})
\end{cases}
\right. 
\end{split}
\label{equAUCROC}
\end{equation}
\newline \textbf{AUC-PR~\cite{10.1145/1143844.1143874}: } The Area Under the Precision-Recall curve (AUC-PR) is defined as the area under the curve corresponding to the Recall on the x-axis and Precision on the y-axis when we vary the anomaly score threshold. 
As before, the area under the curve can be calculated using the trapezoidal rule, defined as follows:

{\footnotesize
\begin{equation}
\begin{split}
&AUC\text{-}PR = \frac{1}{2}\sum_{k=1}^{N} \Delta^{k}_{Precision}*\Delta^{k}_{Recall}\\
&\text{with: } \left.
\begin{cases}
\Delta^{k}_{Recall} &= Recall(Th_{k})-Recall(Th_{k-1})\\
\Delta^{k}_{Precision} &= Precision(Th_{k-1})+Precision(Th_{k})
\end{cases}
\right. 
\end{split}
%\vspace{-0.1cm}
\label{equAUCPR}
\end{equation}
}

\noindent As discussed in~\cite{10.1145/1143844.1143874}, linear interpolation in PR space may result in an overly optimistic estimate of performance. Therefore, we adopt an alternative interpolation method, Stepwise Interpolation, to approximate the area under the curve by calculating the average precision of the PR curve:

%\vspace{-0.2cm}
\begin{equation}
AUC\text{-}PR = \sum_{k=1}^{N} Precision(Th_{k})*\Delta^{k}_{Recall}
%\vspace{-0.2cm}
\end{equation}

\noindent For consistency, we use the above equation in this paper to compute AUC-PR.

\section{Problem motivation and limitations}
\label{sec:problem}

\begin{figure}
 \centering
 \includegraphics[height=13cm,width=\linewidth]{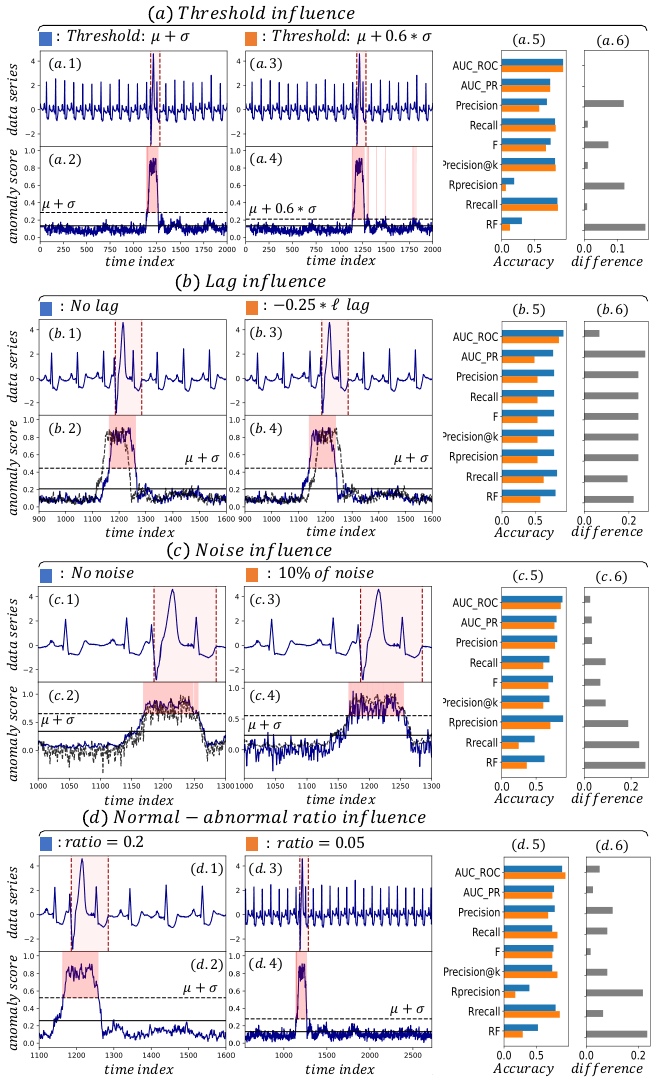}
 %\vspace{-0.7cm}
 \caption{Evaluation measures when we vary the (a) threshold, (b) lag, (c) noise, and (d) normal/abnormal ratio. Example with Isolation Forest methods over a snippet of an ECG time series~\cite{goldberger_physiobank_2000}.}
 \label{fig:limitation_robust}
 \vspace{-0.1cm}
\end{figure}

Having introduced existing measures to assess the quality of range-based anomalies, we now elaborate on their critical limitations.

%\vspace{-0.1cm}
\subsection{Limitations of Threshold-based Measures}
%\vspace{-0.1cm}

The need to threshold the anomaly score severely impacts the accuracy measures. First, Figure~\ref{fig:limitation_robust}(a) depicts an electrocardiogram time series with an arrhythmia in red (Figure~\ref{fig:limitation_robust}(a.1)) and the corresponding anomaly score computed with Isolation Forest~\cite{liu_isolation_2008} (Figure~\ref{fig:limitation_robust}(a.2)) for one threshold equal to $\mu(score) + \sigma(score)$ and for another threshold $\mu(score) + 0.6*\sigma(score)$ (Figures~\ref{fig:limitation_robust}(a.3) and (a.4)). We compute the different accuracy measures for the first threshold (blue bars in Figure~\ref{fig:limitation_robust}(a.5)) and the second threshold (orange bars in Figure~\ref{fig:limitation_robust}(a.5)) and their differences (Figure~\ref{fig:limitation_robust}(a.6)). We observe that the threshold choice has a substantial impact on Precision, Rprecision, F and RF scores. On the contrary, the threshold-independent measures (i.e., measures computing all possible thresholds), namely, AUC-ROC and AUC-PR, show a clear advantage.

Overall, the threshold choice depends on the application and the type of input time series. Setting the threshold automatically is hard and almost impossible when we compare different categories of AD methods across heterogeneous benchmarks. To illustrate this point, we consider two transformations of the anomaly score that correspond to practical cases we observed (e.g., different methods introduce different lag and noise levels to the anomaly score).
\newline \textbf{Influence of Noise: } Some AD methods applied to some specific time series might result in a noisy anomaly score. In addition, due to manufacturing issues or external causes, a sensor can inject noise into the time series, which then propagates on the anomaly score. Figure~\ref{fig:limitation_robust}(c) depicts two cases: the first corresponds to an anomaly score without any noise (Figure~\ref{fig:limitation_robust}(c.2)). The second corresponds to an anomaly score with noise (Figure~\ref{fig:limitation_robust}(c.2)). We applied on both cases the same threshold $\mu(score) + \sigma(score)$. We observe in Figure~\ref{fig:limitation_robust}(c.6) that most of the threshold-based measures are strongly impacted by noise. This is caused by the fact that the score fluctuates around the threshold, making threshold-based measures less robust to noise. On the contrary, AUC-ROC and AUC-PR are much less influenced by noise, returning approximately the same value.
\newline \textbf{Influence of Normal/Abnormal Ratio: } Depending on the domain and the task, the number of anomalies and, consequently, the normal/abnormal ratio changes drastically. A variation in this ratio might cause a variation in the threshold, which leads to variations in threshold-based accuracy measure values. This is explained by the fact that if an anomaly score detects the anomalies correctly, the standard deviation of that score will be higher for a time series with more anomalies. Figure~\ref{fig:limitation_robust}(d) depicts two cases: one time series snippet with a $0.2$ ratio (Figure~\ref{fig:limitation_robust}(d.2)) and one time series snippet with a $0.05$ ratio (Figure~\ref{fig:limitation_robust}(d.4)). We observe that this change implies a larger variation for several threshold-based measures. Thus, the latter confirms the limitations and the non-robustness of threshold-based measures to the anomaly cardinality ratio.

%\vspace{-0.2cm}
\subsection{Limitations of Point-based Measures}
%\vspace{-0.1cm}

In the previous section, we illustrated the limitations of threshold-based measures. By design and because of their independence from the threshold choice, AUC-ROC and AUC-PR measures are robust to those limitations. However, those measures are designed for point-based outliers. Each point is considered independently and the detection of each point has an equivalent contribution to AUC. In contrast, we need to consider two factors, the range detection and the existence detection, for the subsequence AD problem.

The range detection has the same methodology as point detection. We prefer that the algorithm detects every point in the subsequence anomaly. The existence detection is a loose but crucial estimation for the anomaly subsequence detector: detecting a tiny segment of one subsequence outlier is still of great value. 
\newline \textbf{Mismatch between the anomaly score and labels: } Compared to point-based AD, range-based AD encourages accurate capturing of each subsequence anomaly, but the existence detection is good enough to be partially rewarded. Two other reasons support the application of this coarse estimation. 

\begin{figure}[tb]
 \centering
 \includegraphics[height=5.5cm,width=\linewidth]{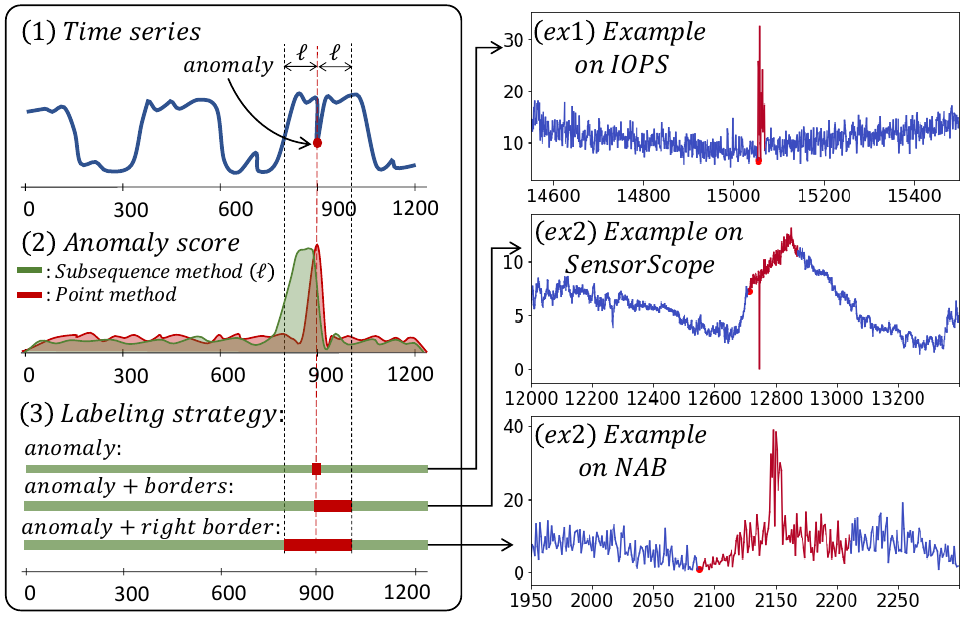}
 %\vspace{-0.7cm}
 \caption{Influence of the anomaly detection method score (2) and labeling strategy (3), illustrated with three examples.}
 \label{fig:influence_labeling}
 %\vspace{-0.2cm}
\end{figure}

First, there is no consistent labeling tradition among datasets. Some may label the whole period as an anomaly if this period does not repeat the typical pattern, while others may only mark a partial period. Figure~\ref{fig:influence_labeling}(3) depicts different labeling strategies. Figures~\ref{fig:influence_labeling}(ex1), (ex2), and (ex3) depict three real examples corresponding to three different labeling strategies that we observed in existing datasets (see Table~\ref{table:charac}). Even if we specify that each period should share the same label, the next question is how to define the starting and end points of a period. Given accurate starting or end points, it is also challenging to label a small segment in one period. Unlike a point outlier, which is an evident deviation from the trend line of the time series, range-based anomalies may not have outrageous values. This difficulty of labeling is inevitable when we assign the discrete labels to a continuous time series. There may be a transition region between the two statuses, but we have to decide on a discontinuous jumping point artificially.

Second, many algorithms, for instance, LOF~\cite{breunig_lof_2000} and iForest~\cite{liu_isolation_2008}, would first apply a sliding window to convert a 1-D time series to a set of high-dimensional data points. We denote the original time series as ($T_1$, $T_2$, $\dots$, $T_n$), and suppose the length of window is $\ell$, then the converted data set is $\{(T_i, \dots, T_{i+\ell-1})|i \in \{1, \dots T-\ell+1\}\}$. The label of point $T_k$ in the time series is defined as the label of high-dimensional point ($T_{k-\ell/2}$, $\dots$, $T_{k+\ell/2-1}$) in the transformed dataset. The conversion from a time series to a dataset has one consequence: every dimension in the high-dimensional point is equally important. So, an abnormal value at the middle or end of this point has the same ability to make it an outlier in the high-dimensional space. Usually, if the sliding window covers more anomaly points, a good algorithm should give a higher anomaly score to the converted data point. However, there are some exceptions, such as that one abnormal value at the beginning or the end of sliding windows is enough to make the converted point an outlier. 
To summarize, an anomaly subsequence $(T_s, \dots, T_e)$ may induce a high anomaly score for the range [$T_{s-\ell/2}, T_{e+\ell/2}$]. A perfect result is that the peak of the anomaly score is slightly broader than the whole abnormal region. The latter is illustrated in Figure~\ref{fig:influence_labeling}(2). However, the anomaly score is not perfect. A high score may be assigned at the range [$T_{s-\ell/2}, T_s$], which fails to reveal the entire range of the outlier but succeeds in indicating the starting region. AUC-based measures will give a low value since there is no overlap between the peak and the outlier.
\newline \textbf{Overall Limitations due to Lag: } A lag can be injected into the anomaly score depending on the choice of AD methods. Overall, such a lag may also exist due to the approximation made during the labeling phase. As illustrated in Figure~\ref{fig:limitation_robust}(b), such a lag (even small) has a substantial impact on \textit{all} existing evaluation measures. For example, in Figure~\ref{fig:limitation_robust}(b) AUC-PR fluctuates between $0.75$ and $0.50$ for a lag of $0.25$ of the labeled section length. Among all measures, only the AUC-ROC measure demonstrates to be less sensitive to such lag (as well as noise and normal/abnormal ratio).

%\vspace{-0.2cm}
\subsection{Problem Definition}
%\vspace{-0.1cm}

In summary, our goal is to develop a new anomaly score threshold-independent evaluation measure based on the robust principles of AUC. A promising direction is an extension of AUC for the range-based AD with the following desired properties:

\noindent{\bf Robust to Lag}: Two similar anomaly scores with a slight lag difference should return approximately the same accuracy measures. For example, a high anomaly score near the border of the anomaly should be rewarded as close as a high anomaly score in the middle of the range-based anomaly.
	%\item 

\noindent{\bf Robust to Noise}: Two similar anomaly scores with and without noise should return similar accuracy.
	%\item 

\noindent{\bf Robust to the Anomaly Cardinality Ratio}: This ratio should not impact the accuracy measures.
	%\item 

\noindent{\bf High Separability between Accurate and Inaccurate Methods}: The accuracy measure should well separate accurate from inaccurate methods.
	%\item 

\noindent{\bf Consistent}: Finally, an appropriate accuracy measure should produce consistent scores for similar time series (i.e., rank different anomaly detection methods in a consistent manner).
%\end{itemize}

\noindent Next, we present new accuracy measures to satisfy these properties.

\begin{figure*}
  \centering
  \includegraphics[width=\linewidth]{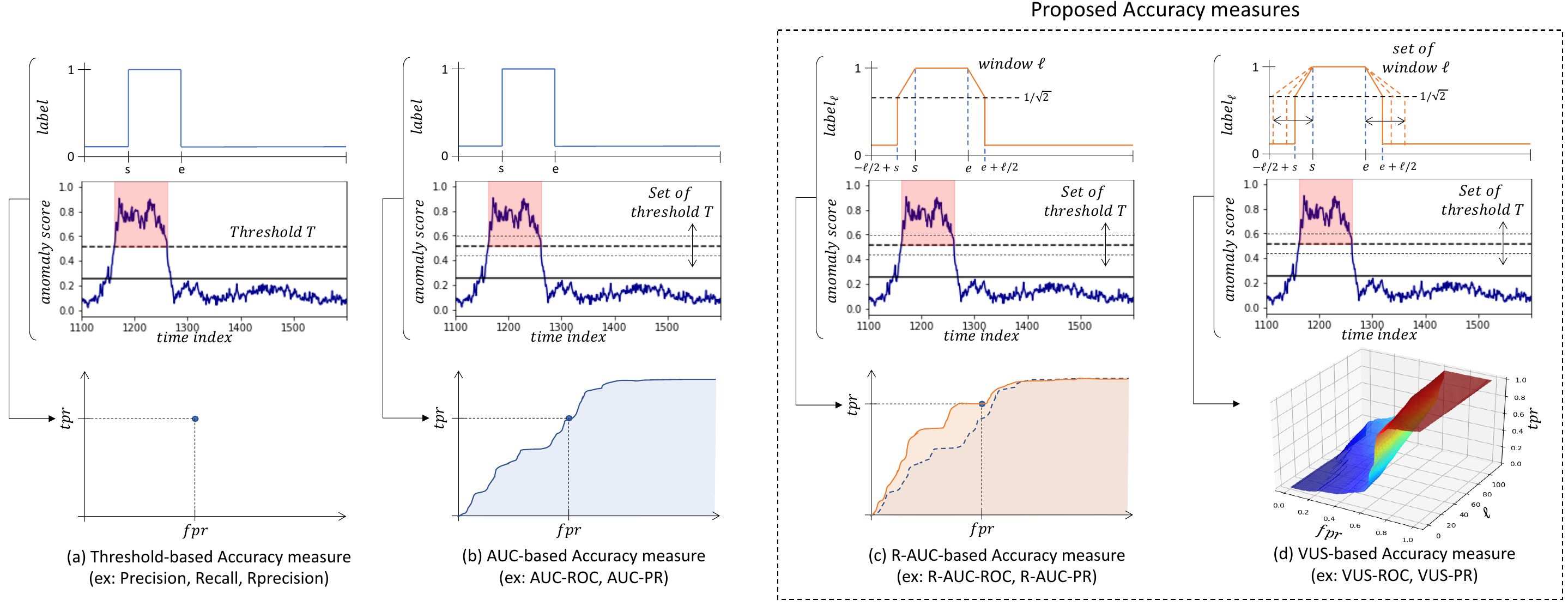}
  %\vspace{-0.7cm}
  \caption{Illustration of previous quality measures compared to our proposed measures. By varying the buffer window, VUS constructs a surface of TPR, FPR, and window. The volume under the surface is a measure of AUC for various windows. }
  \label{fig:auc_volume}
  %\vspace{-0.1cm}
\end{figure*}

\begin{figure}
  \centering
  \includegraphics[width=\linewidth]{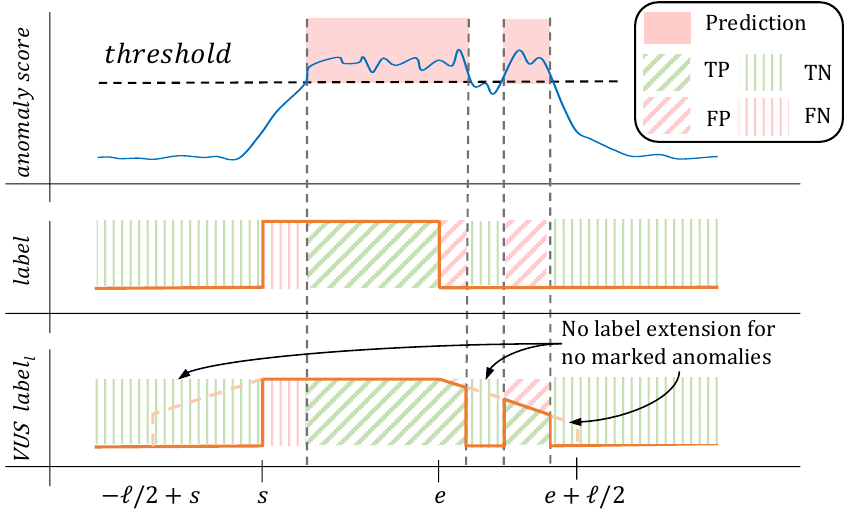}
  %\vspace{-0.7cm}
  \caption{\commentRed{Illustration of proposed label extension strategy.}}
  \label{fig:label_extension}
  %\vspace{-0.1cm}
\end{figure}

%\vspace{-0.1cm}
\section{Our Measures: Range-AUC and VUS}
%\vspace{-0.1cm}

We first present new range-based extensions for ROC and PR curves by introducing a new continuous label to enable more flexibility in measuring detected anomaly ranges. We then present the Volume Under the Surface (VUS) for ROC and PR curves. VUS extends the mathematical model of Range-AUC measures by varying the buffer length. \commentRed{An alternative solution is to learn the necessary parameters and thresholds. However, such a solution works only under supervised settings and may impact the generalizability to new datasets. For the specific case of unsupervised learning, the threshold selection can only be achieved using statistical heuristics. The most common strategy to set the threshold unsupervisely is to set it to $\mu(S_T) + \alpha*\sigma(S_T)$, with $\alpha=3$~\cite{statisticaloutliers}. We will use this strategy when comparing our proposed measures to threshold-based measures.}

%\vspace{-0.2cm}
\subsection{Range-AUC-ROC and Range-AUC-PR}
\label{sec:range-auc}
%\vspace{-0.1cm}

To compute the ROC curve and PR curve for a subsequence, we need to extend to definitions of TPR, FPR, and Precision. 
The first step is to add a buffer region at the boundary of outliers. The idea is that there should be a transition region between the normal and abnormal subsequences to accommodate the false tolerance of labeling in the ground truth (as discussed, this is unavoidable due to the mapping of discrete data to continuous time series). An extra benefit is that this buffer will give credit to the high anomaly score in the vicinity of the outlier boundary, which is what we expected with the application of a sliding window originally. 

Figure ~\ref{fig:auc_volume}(b) shows the original binary labels (in blue), and Figure ~\ref{fig:auc_volume}(c) the new label with buffer region (in orange). By default, the width of the buffer region at each side is half of the period $w$ of the time series (the period is an intrinsic characteristic of the time series). Differently, this parameter can be set into the average length of anomaly sizes or can be set to a desired value by the user.

The traditional binary label is extended to a continuous value. Formally, for a given buffer length $\ell$, the positions $s,e \in [0,|label|]$ the beginning and end indexes of a labeled anomaly (i.e., sections of continuous $1$ in $label$), we define the continuous $label_r$ as follows:
%\vspace{-0.1cm}
\begin{equation}
\footnotesize{
\begin{split}
&\forall i \in [0,|label|], \quad label_{\ell i} \\
& = \begin{cases}
\bigg(1-\frac{|s-i|}{\ell}\bigg)^{\frac{1}{2}}, & \text{if } s-\frac{\ell}{2} \leq i < s \text{ and } {pred}_i = 1, \\
1, & \text{if } s \leq i < e, \\
\bigg(1-\frac{|e-i|}{\ell}\bigg)^{\frac{1}{2}}, & \text{if } e \leq i < e+\frac{\ell}{2} \text{ and } {pred}_i = 1, \\
0, & \text{else}.
\end{cases}
\end{split}
\label{label_equation}
}
%\vspace{-0.1cm}
\end{equation}

\commentRed{
\noindent Specifically, if no predicted anomaly exists within the extended buffer region, we set ${label_{\ell}}_i$ to $0$ to prevent unnecessary false negatives caused by excessive label extension, as illustrated in Figure~\ref{fig:label_extension}.
}
\noindent When the buffer regions of two discontinuous outliers overlap, the label will be the superposition of these two orange curves with one as the maximum value. Using this new continuous label, one can compute $TP_\ell$, $FP_\ell$, $TN_\ell$ and $FN_\ell$ similarly as follows:
%\vspace{-0.2cm}
\begin{equation}
{\small
\begin{split}
&TP_{\ell} = label_{\ell}^\top \cdot pred &FP_{\ell} = (I- label_{\ell})^\top \cdot pred \\
&TN_{\ell} = (I- label_{\ell})^\top \cdot (I-pred) &FN_{\ell} = label_{\ell}^\top \cdot (I-pred) \\
\end{split}
} % font size
%\vspace{-0.2cm}
\end{equation}
\noindent The total number of positive points P in this case naively should be $P_{{\ell}_0} = TP_{\ell}+ FN_{\ell} = label_{\ell}^\top \cdot I$. Here, we define it as:
%\vspace{-0.2cm}
\begin{equation}
%\begin{split}
P_{\ell} = (label+label_{\ell})^\top \cdot \frac{I}{2} \text{, } N_{\ell} = |label_{\ell}|-P_{\ell}
%\end{split}
%\vspace{-0.2cm}
\end{equation}
\noindent The reason is twofold. When the length of the outlier is several periods, $P_{{\ell}_0}$ and $P_{\ell}$ are similar because the ratio of the buffer region to the whole anomaly region is small. When the length of the outlier is only half-period, the size of the buffer region is nearly two times the original abnormal region. In other words, to pursue false tolerance, the relative change we make to the ground truth is too significant. We use the average of $label$ and $label_{\ell}$ to limit this change.

We finally generalize the point-based $Recall$, $Precision$, and $FPR$ to the range-based variants. Formally, following the definition of $R$ and $P$ as the set of anomalies range and detected predicted anomaly range (see Section~\ref{acc_measure}), we define $TPR_{\ell}$, $FPR_{\ell}$, and $Precision_{\ell}$:
%\vspace{-0.2cm}
\begin{equation}
{\small
\begin{split}
TPR_\ell&=Recall_{\ell}=\frac{TP_{\ell}}{P_{\ell}}*\sum_{R_i \in R} \frac{ExistenceR(R_i,P)}{|R|} \\
FPR_{\ell}&=\frac{FP_{\ell}}{N_{\ell}} \text{, } Precision_{\ell}=\frac{TP_{\ell}}{TP_{\ell}+FP_{\ell}} \\
\end{split}
} % font size
%\vspace{-0.2cm}
\label{eqution_constant}
\end{equation}
\noindent Note that $TPR_r=Recall_r$. Moreover, for the recall computation, we incorporate the idea of Existence Reward \cite{tatbul_precision_2018}, which is the ratio of the number of detected subsequence outliers to the total number of subsequence outliers. However, consistent with their work \cite{tatbul_precision_2018}, we do not include the Existence ratio in the definition of range-precision. We can then compute R-AUC-ROC and R-AUC-PR using Equation~\ref{equAUCROC} and Equation~\ref{equAUCPR}.
\newline \textbf{Relation between Range-ROC and Range-PR: } PR curve is a supplement to the ROC curve. In a highly unbalanced dataset, because the number of positive points is too small, at the same level of FPR, it is easy to have a high TPR (or $TPR_{\ell}$) at the cost of low precision.  There are deep connections between ROC and PR \cite{10.1145/1143844.1143874}. First, ROC and PR have one-to-one mapping for a given dataset because the confusion matrix is uniquely determined given TPR and FPR. This relation is broken for the range method because we include an extra Existence factor for range-TPR. Therefore, the confusion matrix cannot be decided in the range-ROC space. Secondly, for a point-based version, if one ROC curve \textit{dominates} another ROC curve, its corresponding PR curve would also dominate another one. Here, dominate means the curve is always higher or equal to another one. Because of the Existence factor, this rule is also lifted for the range definition. This is true only if both of the methods have the same existence ratio. However, this is not always guaranteed. Finally, a maximized AUC does not necessarily correspond to a maximized AP. This holds for the range definition.

\subsection{VUS: Volume Under the Surface}
\label{sec:vus}

Our range-AUC family of measures chooses the width of the buffer region to be half of a subsequence length $\ell$ of the time series. Such buffer length can be either set based on the knowledge of an expert (e.g., the usual size of arrhythmia in an electrocardiogram) or set automatically using the time series's period. \commentRed{The latter can be computed using different strategies: (I) using the Fourier transform to identify the most relevant period of the time series, or (ii) computing the cross-correlation and retrieving the lag value (i.e., subsequence length) that locally maximize the correlation. In practice, we observe that computing the cross-correlation of a time series and selecting the length corresponding to the first local maximal is accurate. In addition, the latter allows users to consider the shortest period length, significantly limiting the execution time of most of the AD methods and the range-AUC measures.} 

Since the period is an intrinsic property of the time series, we can compare various algorithms on the same basis. However, a different approach may get a slightly different period. In addition, there are multi-period time series. So other groups may get different range-AUC because of the difference in the period. As a matter of fact, the parameter $\ell$, if not well set, can strongly influence range-AUC measures. To eliminate this influence, we introduce two generalizations of range-AUC measures.

The solution is to compute ROC and PR curves for different buffer lengths from 0 to $\ell$ as shown in Figure~\ref{fig:auc_volume}(d). Therefore, ROC and PR curves become a surface in a three-dimensional space. Then, the overall accuracy measure corresponds to the Volume Under the Surface (VUS) for either the ROC surface (VUS-ROC) or PR surface (VUS-PR). As the R-AUC-ROC and R-AUC-PR are measures independent of the threshold on the anomaly score, the VUS-ROC and VUS-PR are independent of both the threshold and buffer length. Formally, given $Th=[Th_0,Th_1,...Th_N]$ with $0=Th_0<Th_1<...<Th_N=1$, and $\mathcal{L}=[\ell_0,\ell_1,...,\ell_L]$ with $0=\ell_0<\ell_1< ... < \ell_L = \ell$, we have:
%\vspace{-0.1cm}
\begin{equation}
\footnotesize{
\begin{split}
&VUS\text{-}ROC = \frac{1}{4}\sum_{w=1}^{L} \sum_{k=1}^{N} \Delta^{(k,w)} * \Delta^{w} \text{, with: }\\
&\left.
\begin{cases}
\Delta^{(k,w)} &= \Delta^{k}_{TPR_{\ell_w}}*\Delta^{k}_{FPR_{\ell_w}}+\Delta^{k}_{TPR_{\ell_{w-1}}}*\Delta^{k}_{FPR_{\ell_{w-1}}} \\
\Delta^{k}_{FPR_{\ell_w}} &= FPR_{\ell_w}(Th_{k})-FPR_{\ell_w}(Th_{k-1}) \\
\Delta^{k}_{TPR_{\ell_w}} &= TPR_{\ell_w}(Th_{k-1})+TPR_{\ell_w}(Th_{k}) \\
\Delta^{w} &= |\ell_w - \ell_{w-1}|
\end{cases}
\right. 
\end{split}
\label{equVUSROC}
}
%\vspace{-0.1cm}
\end{equation}

%\vspace{-0.1cm}
\begin{equation}
\footnotesize{
\begin{split}
&VUS\text{-}PR = \frac{1}{2}\sum_{w=1}^{L} \sum_{k=1}^{N} \Delta^{(k,w)} * \Delta^{w} \text{, with: }\\
&\left.
\begin{cases}
\Delta^{(k,w)} &= {Precision_{\ell_w}(Th_k)}*\Delta^{k}_{Re_{\ell_w}}\\  &\quad+{Precision_{\ell_{w-1}}(Th_k)}*\Delta^{k}_{Re_{\ell_{w-1}}} \\
\Delta^{k}_{Re_{\ell_w}} &= Recall_{\ell_w}(Th_{k})-Recall_{\ell_w}(Th_{k-1}) \\
% \Delta^{k}_{Pr_{\ell_w}} &= Precision_{\ell_w}(Th_{k-1})+Precision_{\ell_w}(Th_{k}) \\
\Delta^{w} &= |\ell_w - \ell_{w-1}|
\end{cases}
\right. 
\end{split}
\label{equVUSPR}
}
%\vspace{-0.1cm}
\end{equation} 

From the above equations, VUS measures are more expensive to compute than range-AUC measures.
Thus, the application of VUS versus range-AUC depends on our knowledge of which buffer length to set. If one user knows which would be the most appropriate buffer length, range-AUC-based measures are preferable compared to VUS-based measures.
However, if there exists an uncertainty on $\ell$, then setting a range and using VUS increases the flexibility of the usage and the robustness of the evaluation. Finally, more parameters than $\ell$ can be included in VUS-based measures. If, in addition to $\ell$, there is a need to define a range for another parameter (such as the normal model length $\ell_{N_M}$ of NormA), the two-dimensional surface is transformed into a three-dimensional hyper-surface. In general, for $P$ parameters, the value is the volume under a $|P|-1$ hyper-surface.

\subsubsection{{\bf Complexity Analysis}}\hfill\\

This section analyzes the complexity of the VUS-based measures. 
We take into account both computation time, and memory usage.

\begin{algorithm}[tb]
{\small
    \caption{\textbf{VUS algorithm}}\label{alg:VUS}
    \label{alg:vus}
    \SetKwInOut{Input}{input}
    \SetKwInOut{Output}{output}
    \Input{Original Labels $label$, anomaly score $S_{T}$, maximum Buffer Length $L$, thresholds $N$}
    \Output{VUS\_ROC, VUS\_PR}
    \BlankLine
    $Th$ $\leftarrow$ $Thresholds(N)$\;
    $\mathcal{L}$ $\leftarrow$ $Buffer\_Lengths(L)$\;
    AUC $\leftarrow$ [],
    AP $\leftarrow$ []\;
    \tcp{Iterate through the buffer Lengths}
    \ForEach{$\ell \in \mathcal{L}$ }
    {
        $Create$ $label_\ell$ from $label$ and $\ell$\;
        $seq$= $Anomaly\_Index(label_\ell)$\;
        $list\_TPR_{\ell}$ $\leftarrow$ [],
        $list\_FPR_{\ell}$ $\leftarrow$ [],
        $list\_Prec_{\ell}$ $\leftarrow$ []\;
        \tcp{Iterate through the thresholds}
        \ForEach{$threshold \in Th$}
        {   
            $pred$ $\leftarrow$ $S_{T}>threshold$\;
            $Change$ $label_\ell$ to $label_\ell^{thres}$ based on $pred$\;
            $product$ $\leftarrow$ $label_\ell^{thres}*pred$\;
            $SumPred$ $\leftarrow$ $\sum_{p\in pred} p$\;
            $SumLabel$ $\leftarrow$ $\sum_{p\in label_\ell^{thres}} p$\;
            $TP_\ell$ $\leftarrow$ 0\;
            \ForEach{$seg \in seq_L$}
            {
                $TP_\ell$ $\leftarrow$ $TP_\ell$ + $\sum_{p\in product[seg[0]:(seg[1]+1)]}p$
            }
            $Compute$ $FP_\ell$, $P_\ell$, $N_\ell$\ from $TP_\ell$, $SumPred$, $SumLabel$\;% $\leftarrow$ $\sum_{p\in product}p$\;
            %$FP_\ell$ $\leftarrow$ $\sum_{p\in product}p$\;
            %$P_\ell$ $\leftarrow$ $\sum_{l_1,l_2\in label,label_\ell} \frac{(l_1+l_2)}{2}$
            
            $Existence_{seq}$ $\leftarrow$ 0\;
            \tcp{Iterate through the anomalies}
            \ForEach{$seg \in seq$}
            {
                \If{$\sum_{p\in product[seg[0]:(seg[1]+1)]}p>0$}
                {
                    $Existence_{seq}$ $\leftarrow$ $Existence_{seq}$ + 1
                }
                $Existence$ $\leftarrow$ $\frac{Existence_{seq}}{|seq|}$
                  
            }
            $Append$ $\frac{TP_\ell*Existence}{P_\ell}$ in $list\_TPR_{\ell}$\;
            $Append$ $\frac{FP_\ell}{N_\ell}$ in $list\_FPR_{\ell}$\;
            $Append$ $\frac{TP_\ell}{TP_\ell+FP_\ell}$ in $list\_Prec_{\ell}$\;
        }
        $Compute$ AUC\_r, AP\_r $from$ $list\_TPR_{\ell}$,$list\_FPR_{\ell}$ and $list\_Prec_{\ell}$\;
        $Append$ AUC\_r, AP\_r $in$ AUC, AP\;
    }
    \tcp{Avg. across thresholds and buffer lengths}
    VUS\_ROC $\leftarrow$ $\frac{\sum_{a\in AUC}a}{|\mathcal{L}|}$,
    VUS\_PR $\leftarrow$ $\frac{\sum_{a\in AP}a}{|\mathcal{L}|}$\;
 % font size
 }
\end{algorithm}

{\bf [Time Complexity]}
The time complexity of VUS (both VUS-ROC and VUS-PR) is determined by varying two parameters, namely the buffer length $\ell \in \mathcal{L}$ and the number of thresholds $N$.
This is further illustrated in Algorithm~\ref{alg:vus}, which breaks down the computation steps. 
It comprises a nested loop that demonstrates the variation of the parameters buffer length \commentRed{($\mathcal{L}$ lengths in total)} and number of thresholds \commentRed{($N$ thresholds in total)}. \commentRed{Therefore, VUS complexity is quadratic to $N$ and $L$. Then, for each threshold and length (inside the loop) the computational complexity is $O(\alpha \ell_a + T_1 + T_2)$}, where $\alpha$ is the number of anomalies, $\ell_a$ refers to the mean length of anomalies, and $T_1, T_2$ refer to computations in the order of length of the time series $T$ involved in the anomaly detection. 
There is a distinction between $T_1$ and $T_2$ because their practical implementations are optimized to different extents, producing very different execution times. 
Here, $O(T_1)$ is the complexity of the calculations involving the entire time series, such as $pred$ (i.e., a boolean sequence indicating if a point of the anomaly score $S_T$ is above a given threshold), and $label_\ell$ (i.e., the modified label sequence with buffer regions). $O(T_2)$ refers to the complexity of the computation of $product$, $TP_\ell$, $FP_\ell$, $P_\ell$, and $N_\ell$, which has a cost of $|T|$, but is less optimized than the previously described computation. 
Moreover, $\alpha \ell_a$ corresponds to the computation of $Existence$. Thus, the total complexity of the algorithm is $O(NL(\alpha \ell_a+T_1+T_2))$. 
In practice, $\alpha \ell_a$ is negligible compared to $T_1$ or $T_2$, and VUS complexity can be approximated to $O(NL(T_1+T_2))$.

{\bf [Space Complexity]}
The space complexity can be obtained from the pseudo-code in Algorithm~\ref{alg:vus}. 
The computation of VUS-ROC and VUS-PR is performed by iterating over the set of buffer lengths ($\mathcal{L}$) and the set of thresholds ($N$). 
Thus, the space complexity of VUS is $O(NL)$.

\subsection{A faster Implementation of VUS}
\label{sec:fasterimpl}

\begin{figure}[tb]
  \centering
  \includegraphics[width=\linewidth]{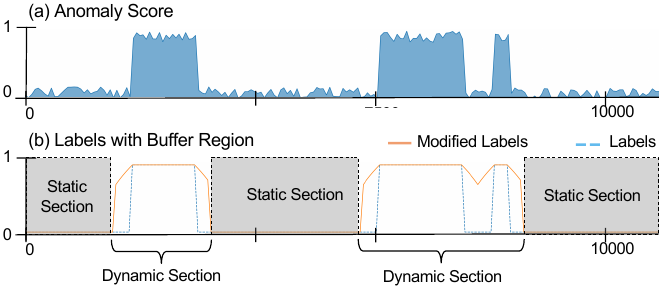}
  %\vspace*{-0.5cm}
  \caption{Synthetic illustration of an anomaly score (a) and labels (b) of a given time series. We differentiate \textbf{static sections} that are invariant to the change of threshold and buffer length, and \textbf{dynamic sections} that have an impact on the accuracy.}
  \label{fig:static_dyn}
\end{figure}

As theoretically explained in the previous section, VUS's computation heavily depends on the time series length, as well as on the number of buffer lengths considered. In this section, we propose a novel implementation that significantly reduces the theoretical computation of the VUS measures.

\begin{algorithm}[tb]
{\small
    \caption{\textbf{\textbf{VUS}$_{opt}$}}\label{alg:VUS_opt}
    \SetKwInOut{Input}{input}
    \SetKwInOut{Output}{output}
    \Input{Original Labels $T$, anomaly score $S_{T}$, maximum Buffer Length $L$, thresholds $N$}
    \Output{VUS-ROC, VUS-PR}
    \BlankLine
    $Th$ $\leftarrow$ $Thresholds(N)$,
    $\mathcal{L}$ $\leftarrow$ $Buffer\_Lengths(L)$\;
    $Create$ $label_L$ from $label$ and $L$\;
    \tcp{Extract anomalies positions for buffer length L}
    $seq_L$ $\leftarrow$ $Anomaly\_Index(label_L)$\;
    $AUC$ $\leftarrow$ [], 
    $AP$ $\leftarrow$ []\;
    \tcp{Static Part}
    \tcp{Iterate through thresholds only}
    \ForEach{$threshold \in Th$}
    {\label{line_vus:static_b}
        $pred$ $\leftarrow$ $S_{T}>threshold$\;
        $SumPred$ $\leftarrow$ $\sum_{p\in pred} p$\;
    }\label{line_vus:static_e}
    \tcp{Dynamic Part}
    \tcp{Iterate through the buffer Lengths}
    \ForEach{$\ell \in \mathcal{L}$ }
    {\label{line_vus:dyn_b}
        $Create$ $label_\ell$ from $label$ and $\ell$\;
        $seq$= $Anomaly\_Index(label_\ell)$\;
        $l\_TPR_{\ell}$ $\leftarrow$ [], 
        $l\_FPR_{\ell}$ $\leftarrow$ [], 
        $l\_Prec_{\ell}$ $\leftarrow$ []\;
        \tcp{Iterate through the thresholds}
        \ForEach{$threshold \in Th$}
        {   
            $pred$ $\leftarrow$ $S_{T}>threshold$\;
            $Change$ $label_\ell$ to $label_\ell^{thres}$ based on $pred$\;
            $product$ $\leftarrow$ $label_\ell^{thres}$*$pred$\;
            $SumLabel$ $\leftarrow$ $\sum_{p\in label_\ell^{thres}} p$\;
            $TP_\ell$ $\leftarrow$ 0\;
            \ForEach{$seg \in seq_L$}
            {
                $TP_\ell$ $\leftarrow$ $TP_\ell$ + $\sum_{p\in product[seg[0]:(seg[1]+1)]}p$
            }
            $Compute$ $FP_\ell$, $P_\ell$, $N_\ell$\ from $TP_\ell$, $SumPred$, $SumLabel$\;
            
            $Existence_{seq}$ $\leftarrow$ 0\;
            \tcp{Iterate through the anomalies}
            \ForEach{$seg \in seq$}
            {
                \If{$\sum_{p\in product[seg[0]:(seg[1]+1)]}p>0$}
                {
                    $Existence_{seq}$ $\leftarrow$ $Existence_{seq}$ + 1
                }
                $Existence$ $\leftarrow$ $\frac{Existence_{seq}}{|seq|}$
                  
            }
            $Append$ $\frac{TP_\ell*Existence}{P_\ell}$ in $l\_TPR_{\ell}$\;
            $Append$ $\frac{FP_\ell}{N_\ell}$ in $l\_FPR_{\ell}$\;
            $Append$ $\frac{TP_\ell}{TP_\ell+FP_\ell}$ in $l\_Prec_{\ell}$\;
        }
        $Compute$ $AUC_r$, $AP_r$ $from$ $l\_TPR_{\ell}$,$l\_FPR_{\ell}$ and $l\_Prec_{\ell}$\;
        $Append$ $AUC_r$, $AP_r$ $in$ $AUC$, $AP$\;
    } \label{line_vus:dyn_e}
    \tcp{Avg. across thresholds and buffer lengths}
    VUS-ROC $\leftarrow$ $\frac{\sum_{a\in AUC}a}{|\mathcal{L}|}$, 
    VUS-PR $\leftarrow$ $\frac{\sum_{a\in AP}a}{|\mathcal{L}|}$\;
 % font size
 }
\end{algorithm}

\begin{algorithm}
{\small
    \caption{\textbf{VUS$_{opt}^{mem}$}}\label{alg:VUS_opt^{mem}}
    \SetKwInOut{Input}{input}
    \SetKwInOut{Output}{output}
    \Input{Original Labels $T$, anomaly score $S_{T}$, maximum Buffer Length $L$, thresholds $N$}
    \Output{VUS-ROC, VUS-PR}
    \BlankLine
    $Th$ $\leftarrow$ $Thresholds(N)$,
    $\mathcal{L}$ $\leftarrow$ $Buffer\_Lengths(L)$\;
    $Create$ $label_L$ from $label$ and $L$\;
    \tcp{Extract anomalies positions for buffer length L}
    $seq_L$ $\leftarrow$ $Anomaly\_Index(label_L)$\;
    $AUC$ $\leftarrow$ [], 
    $AP$ $\leftarrow$ []\; 
    $Pred_{Th}$ $\leftarrow$ []\;
    \tcp{Static Part}
    \tcp{Iterate only through thresholds}
    \ForEach{$threshold \in Th$}
    {
        $pred$ $\leftarrow$ $S_{T}>threshold$\;
        $Pred_{Th}$ $\leftarrow$ Append with $pred$\;
        $SumPred$ $\leftarrow$ $\sum_{p\in pred} p$\;
    }
    \tcp{Dynamic Part}
    \tcp{Iterate through the buffer Lengths}
    \ForEach{$\ell \in \mathcal{L}$ }
    {
        $Create$ $label_\ell$ from $label$ and $\ell$\;
        $seq$= $Anomaly\_Index(label_\ell)$\;
        $l\_TPR_{\ell}$ $\leftarrow$ [],
        $l\_FPR_{\ell}$ $\leftarrow$ [],
        $l\_Prec_{\ell}$ $\leftarrow$ []\;
        \tcp{Iterate through the thresholds}
        count $\leftarrow$ 0\;
        \ForEach{$threshold \in Th$}
        {  
            $Change$ $label_\ell$ to $label_\ell^{thres}$ based on $Pred_{Th}[threshold]$\;
            $product$ $\leftarrow$ $label_\ell^{thres}*Pred_{Th}[threshold]$\;
            $SumLabel$ $\leftarrow$ $\sum_{p\in label_\ell^{thres}} p$\;
            $TP_\ell$ $\leftarrow$ 0\;
            \ForEach{$seg \in seq_L$}
            {
                $TP_\ell$ $\leftarrow$ $TP_\ell$ + $\sum_{p\in product[seg[0]:(seg[1]+1)]}p$
            }
            $Compute$ $FP_\ell$, $P_\ell$, $N_\ell$\ from $TP_\ell$, $SumPred$, $SumLabel$\;
            $Existence_{seq}$ $\leftarrow$ 0\;
            \tcp{Iterate through the anomalies}
            \ForEach{$seg \in seq$}
            {
                \If{$\sum_{p\in product[seg[0]:(seg[1]+1)]}p>0$}
                {
                    $Existence_{seq}$ $\leftarrow$ $Existence_{seq}$ + 1
                }
                $Existence$ $\leftarrow$ $\frac{Existence_{seq}}{|seq|}$
                  
            }
            $Append$ $\frac{TP_\ell*Existence}{P_\ell}$ in $l\_TPR_{\ell}$\;
            $Append$ $\frac{FP_\ell}{N_\ell}$ in $l\_FPR_{\ell}$\;
            $Append$ $\frac{TP_\ell}{TP_\ell+FP_\ell}$ in $l\_Prec_{\ell}$\;   
        }
        $Compute$ AUC\_r, AP\_r $from$ $l\_TPR_{\ell}$,$l\_FPR_{\ell}$ and $l\_Prec_{\ell}$\;
        $Append$ AUC\_r, AP\_r $in$ AUC, AP\;
    }
    \tcp{Avg. across thresholds and buffer lengths}
    VUS\_ROC $\leftarrow$ $\frac{\sum_{a\in AUC}a}{|\mathcal{L}|}$,
    VUS\_PR $\leftarrow$ $\frac{\sum_{a\in AP}a}{|\mathcal{L}|}$\;
 % font size
 }
\end{algorithm}

\subsubsection{Dynamic versus Static sections}\hfill\\

The variations of thresholds and buffer length affect the modified labels (i.e., $label_\ell$) and $pred$, which cause changes in the values of True and False Positive Rates ($TPR$ and $FPR$). 
However, not all sections of the time series are affected by these variations. 
The data points, whose labels are not affected by the change in the buffer length for a given threshold, have the same $TPR$ and $FPR$ (i.e., data points that lie outside the maximum possible buffer length of an anomaly). 

As a result, we can segment the time series into two categories: $Dynamic$ and $Static$. The first category corresponds to sections of the time series containing labels affected by the variation of buffer length. The second category corresponds to sections that are not affected by these changes. Figure~\ref{fig:static_dyn} illustrates this segmentation, enabling us to compute the same measures with significantly reduced computational costs.

\begin{figure}[tb]
  \centering
  \includegraphics[width=\linewidth]{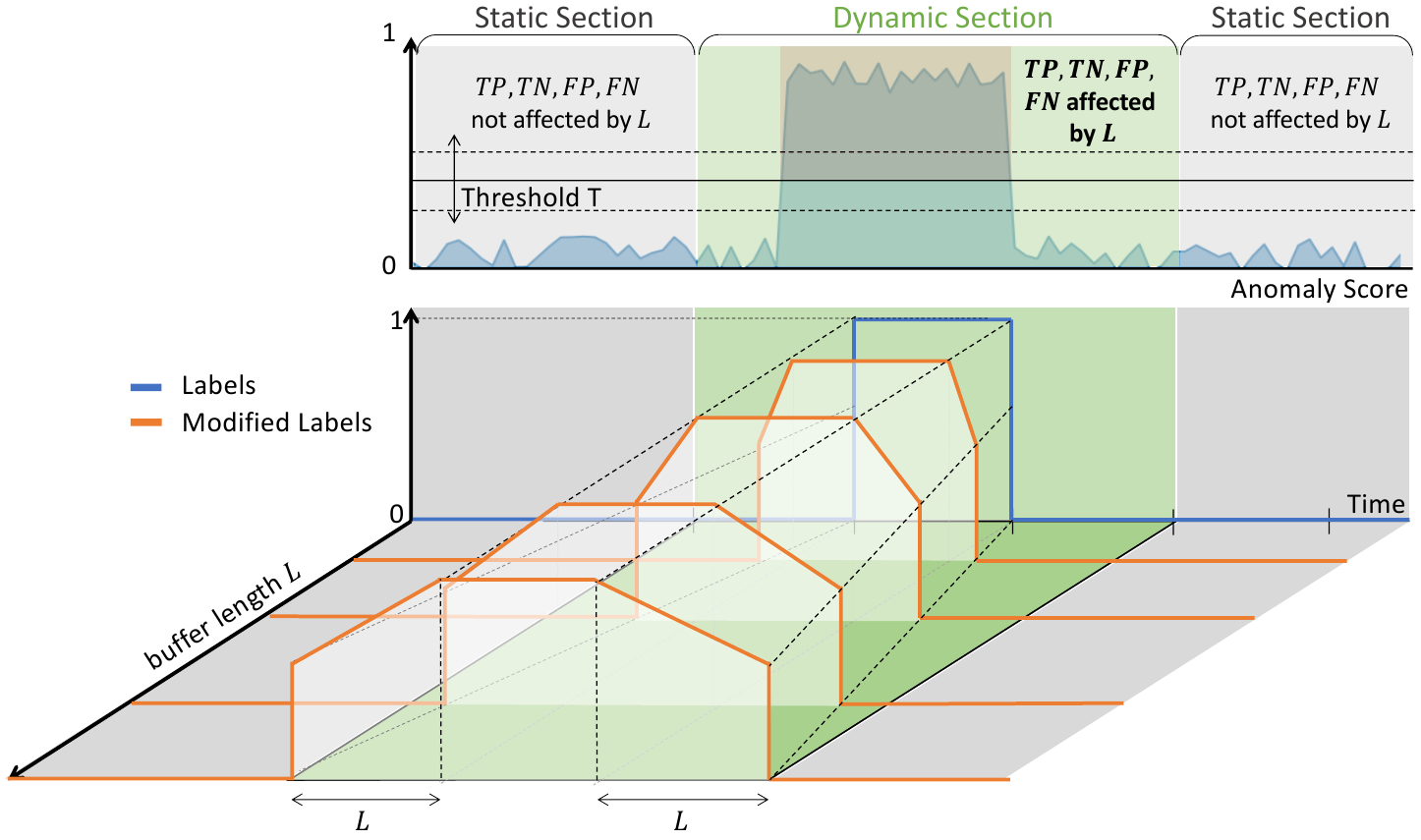}
  %\vspace*{-0.5cm}
  \caption{Synthetic illustration of the labels evolution with $L$. In contrast to dynamic sections (in green), the buffer length has no impact on VUS within the static sections (in grey).}
  \label{fig:static_dyn_2}
\end{figure}

\subsubsection{Algorithmic Implementation}\hfill\\

The optimization described above can be performed following two different strategies:

\begin{itemize}
\item {\bf VUS$_{opt}$}: In this version, we split the time series anomaly scores $S_T$ and $label_\ell$ into static and dynamic sections. We compute the constant required to calculate VUS only once for the static sections, and once for each buffer length and threshold value for the dynamic sections.
\item {\bf VUS$_{opt}^{mem}$}: This version is an extension of the previous, with a code-wise modification that leads to a further decrease in time complexity at the expense of increased space complexity.
Given the large main memory sizes of modern servers (and even desktops and laptops), VUS$_{opt}^{mem}$ represents a very attractive solution in practice.
\end{itemize}

Due to the consideration of splitting data into static and dynamic regions, the implementation has the following advantages:

\begin{itemize}
\item The static split avoids repetitive calculations that would have involved the same values repeatedly in a nested loop (i.e., computing only the accuracy values for each threshold for the static region, since buffer size does not affect static regions).
\item The calculations of $TP$ and $N$ in Equation~\ref{eqution_constant} essentially add up to zero in the above consideration of the static part, and do not need to be computed. 
\item The overall computational time is similar to that of the Range-AUC measures for a relatively small number of anomalies, but even more importantly, it does not increase when the number of anomalies gets significantly larger.
\end{itemize}

The computational steps of $VUS_{opt}$ and $VUS_{opt}^{mem}$ are shown in Algorithm~\ref{alg:VUS_opt} and Algorithm~\ref{alg:VUS_opt^{mem}}, respectively.
These two algorithms are divided into two different sections: (i) the static part in which we compute VUS for sections of the time series without anomalies, and (ii) the dynamic part in which we compute VUS only for the time series sections that contain anomalies.
In the following sections, we analyze in detail the theoretical complexity (space and time).

\noindent{\bf [VUS$_{opt}$ Time and Space Complexity]}: The VUS$_{opt}$ computation is similar to the original VUS computation (cf. Algorithm~\ref{alg:vus}) for the calculations of the dynamic part. 
However, it differs in the static part, as its calculations that involve predictions and labels are unaffected by buffer length. 
The static part computation (Lines~\ref{line_vus:static_b}-\ref{line_vus:static_e}) involves the predictions (according to all possible thresholds in $Th$) and their summation. 
Thus, the complexity for the static sections is $O(N(T_1+T_2))$. 
For the dynamic part (Lines~\ref{line_vus:dyn_b}-\ref{line_vus:dyn_e}), the computations (for each threshold and buffer length) are only performed for the sections containing anomalies (i.e., dynamic sections in Figure~\ref{fig:static_dyn}). Thus, the complexity of the dynamic part computation is $O(\alpha \ell_a)$.
We also have to compute the predictions (score values above a given threshold) for each dynamic section, which have a complexity of $O(T_2)$. 
Thus the total complexity adds up to $O(N(T_1+T_2))+O(NL(\alpha \ell_a+T_2))$.
In addition, the space complexity of the dynamic computation with the nested loop of thresholds and buffer length is $O(NL)$, and $O(N)$ for the static part. Therefore, the overall space complexity of VUS$_{opt}$ is $O(NL)$.

\noindent{\bf [VUS$_{opt}^{mem}$ Time and Space Complexity]}
As shown in Algorithm~\ref{alg:VUS_opt^{mem}}, the complexity of the static sections remains unchanged compared to VUS$_{opt}$. However, the complexity related to the dynamic sections is reduced by reusing the saved predictions calculated in the static part (as illustrated in Figure~\ref{fig:static_dyn_2}, it is not affected by buffer lengths).
This reduces the dynamic complexity to  $O(\alpha \ell_a)$, adding up to a total complexity of $O(N(T_1+T_2)+ NL\alpha \ell_a)$. 
For VUS$_{opt}^{mem}$, similarly to VUS$_{opt}$, the space complexity of the dynamic computation containing the nested loop of thresholds and buffer length is $O(NL)$. However, due to the storage and indexing of predictions, the computations related to the static sections result in a space complexity of $O(NT)$. Thus, the total space complexity of VUS$_{opt}^{mem}$ is $O(N(L+T))$. 
The time and space complexity of all three VUS implementations are listed in Table~\ref{tab:complexity_summary}.

\begin{table}[tb]
    \centering
    \caption{{Space and time complexity of VUS implementations}}
    \scalebox{0.88}{
    \begin{tabular}{|c|c|c|}
    \hline
         Version & Time & Space \\
         \hline
         $VUS$ & $O(NL(\alpha \ell_a+T_1+T_2))$ & $O(NL)$\\ 
 VUS$_{opt}$ & $O(N(T_1+T_2+L(\alpha \ell_a+T_2)))$ &  $O(NL)$\\
 VUS$_{opt}^{mem}$ & $O(N(T_1+T_2+L\alpha \ell_a))$ & $O(N(L+T))$\\ 
 \hline
    \end{tabular}
    }
    \label{tab:complexity_summary}
\end{table}

\section{Experimental Analysis}
\label{sec:exp}
We now describe in detail our experimental analysis. The experimental section is organized as follows:
%\begin{enumerate}[noitemsep,topsep=0pt,parsep=0pt,partopsep=0pt,leftmargin=0.5cm]
%\item 

\noindent In {\bf 
Section~\ref{exp:setup}}, we introduce the datasets and methods to evaluate the previously defined accuracy measures.

%\item
\noindent In {\bf 
Section~\ref{exp:qual}}, we illustrate the limitations of existing measures with some selected qualitative examples.

%\item 
\noindent In {\bf 
Section~\ref{exp:quant}}, we continue by measuring quantitatively the benefits of our proposed measures in terms of {\it robustness} to lag, noise, and normal/abnormal ratio.

%\item 
\noindent In {\bf 
Section~\ref{exp:separability}}, we evaluate the {\it separability} degree of accurate and inaccurate methods, using the existing and our proposed approaches.

%\item
\noindent In {\bf 
Section~\ref{sec:entropy}}, we conduct a {\it consistency} evaluation, in which we analyze the variation of ranks that an AD method can have with an accuracy measures used.

%\item 
\noindent In {\bf 
Section~\ref{sec:exectime}}, we conduct an {\it execution time} evaluation, in which we analyze the impact of different parameters related to the accuracy measures and the time series characteristics. 
We focus especially on the comparison of the different VUS implementations.
%\end{enumerate}

\begin{table}[tb]
\caption{Summary characteristics (averaged per dataset) of the public datasets of TSB-UAD (S.: Size, Ano.: Anomalies, Ab.: Abnormal, Den.: Density)}
\label{table:charac}
%\vspace{-0.2cm}
\footnotesize
\begin{center}
\scalebox{0.82}{
\begin{tabular}{ |r|r|r|r|r|r|} 
 \hline
\textbf{\begin{tabular}[c]{@{}c@{}}Dataset \end{tabular}} & 
\textbf{\begin{tabular}[c]{@{}c@{}}S. \end{tabular}} & 
\textbf{\begin{tabular}[c]{c@{}} Len.\end{tabular}} & 
\textbf{\begin{tabular}[c]{c@{}} \# \\ Ano. \end{tabular}} &
\textbf{\begin{tabular}[c]{c@{}c@{}} \# \\ Ab. \\ Points\end{tabular}} &
\textbf{\begin{tabular}[c]{c@{}c@{}} Ab. \\ Den. \\ (\%)\end{tabular}} \\ \hline
Dodgers \cite{10.1145/1150402.1150428} & 1 & 50400   & 133.0     & 5612.0  &11.14 \\ \hline
SED \cite{doi:10.1177/1475921710395811}& 1 & 100000   & 75.0     & 3750.0  & 3.7\\ \hline
ECG \cite{goldberger_physiobank_2000}   & 52 & 230351  & 195.6     & 15634.0  &6.8 \\ \hline
IOPS \cite{IOPS}   & 58 & 102119  & 46.5     & 2312.3   &2.1 \\ \hline
KDD21 \cite{kdd} & 250 &77415   & 1      & 196.5   &0.56 \\ \hline
MGAB \cite{markus_thill_2020_3762385}   & 10 & 100000  & 10.0     & 200.0   &0.20 \\ \hline
NAB \cite{ahmad_unsupervised_2017}   & 58 & 6301   & 2.0      & 575.5   &8.8 \\ \hline
NASA-M. \cite{10.1145/3449726.3459411}   & 27 & 2730   & 1.33      & 286.3   &11.97 \\ \hline
NASA-S. \cite{10.1145/3449726.3459411}   & 54 & 8066   & 1.26      & 1032.4   &12.39 \\ \hline
SensorS. \cite{YAO20101059}   & 23 & 27038   & 11.2     & 6110.4   &22.5 \\ \hline
YAHOO \cite{yahoo}  & 367 & 1561   & 5.9      & 10.7   &0.70 \\ \hline 
\end{tabular}}
\end{center}
\end{table}

\subsection{Experimental Setup and Settings}
\label{exp:setup}
%\vspace{-0.1cm}

\begin{figure*}[tb]
  \centering
  \includegraphics[width=1\linewidth]{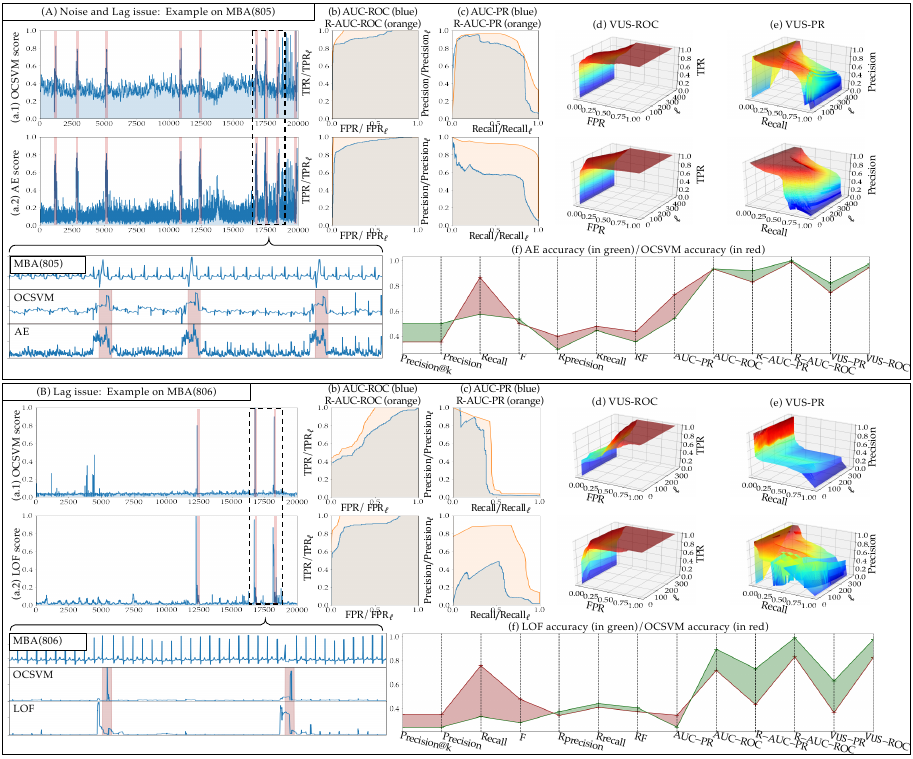}
  %\vspace{-0.7cm}
  \caption{Comparison of evaluation measures (proposed measures illustrated in subplots (b,c,d,e); all others summarized in subplots (f)) on two examples ((A)AE and OCSM applied on MBA(805) and (B) LOF and OCSVM applied on MBA(806)), illustrating the limitations of existing measures for scores with noise or containing a lag. }
  \label{fig:quality}
  %\vspace{-0.1cm}
\end{figure*}

We implemented the experimental scripts in Python 3.8 with the following main dependencies: sklearn 0.23.0, tensorflow 2.3.0, pandas 1.2.5, and networkx 2.6.3. In addition, we used implementations from our TSB-UAD benchmark suite.\footnote{\scriptsize \url{https://www.timeseries.org/TSB-UAD}} For reproducibility purposes, we make our datasets and code available.\footnote{\scriptsize \url{https://www.timeseries.org/VUS}}
\newline \textbf{Datasets: } For our evaluation purposes, we use the public datasets identified in our TSB-UAD benchmark. The latter corresponds to $10$ datasets proposed in the past decades in the literature containing $900$ time series with labeled anomalies. Specifically, each point in every time series is labeled as normal or abnormal. Table~\ref{table:charac} summarizes relevant characteristics of the datasets, including their size, length, and statistics about the anomalies. In more detail:

\begin{itemize}
    \item {\bf SED}~\cite{doi:10.1177/1475921710395811}, from the NASA Rotary Dynamics Laboratory, records disk revolutions measured over several runs (3K rpm speed).
	\item {\bf ECG}~\cite{goldberger_physiobank_2000} is a standard electrocardiogram dataset and the anomalies represent ventricular premature contractions. MBA(14046) is split to $47$ series.
	\item {\bf IOPS}~\cite{IOPS} is a dataset with performance indicators that reflect the scale, quality of web services, and health status of a machine.
	\item {\bf KDD21}~\cite{kdd} is a composite dataset released in a SIGKDD 2021 competition with 250 time series.
	\item {\bf MGAB}~\cite{markus_thill_2020_3762385} is composed of Mackey-Glass time series with non-trivial anomalies. Mackey-Glass data series exhibit chaotic behavior that is difficult for the human eye to distinguish.
	\item {\bf NAB}~\cite{ahmad_unsupervised_2017} is composed of labeled real-world and artificial time series including AWS server metrics, online advertisement clicking rates, real time traffic data, and a collection of Twitter mentions of large publicly-traded companies.
	\item {\bf NASA-SMAP} and {\bf NASA-MSL}~\cite{10.1145/3449726.3459411} are two real spacecraft telemetry data with anomalies from Soil Moisture Active Passive (SMAP) satellite and Curiosity Rover on Mars (MSL).
	\item {\bf SensorScope}~\cite{YAO20101059} is a collection of environmental data, such as temperature, humidity, and solar radiation, collected from a sensor measurement system.
	\item {\bf Yahoo}~\cite{yahoo} is a dataset consisting of real and synthetic time series based on the real production traffic to some of the Yahoo production systems.
\end{itemize}

\textbf{Anomaly Detection Methods: }  For the experimental evaluation, we consider the following baselines. 

\begin{itemize}
\item {\bf Isolation Forest (IForest)}~\cite{liu_isolation_2008} constructs binary trees based on random space splitting. The nodes (subsequences in our specific case) with shorter path lengths to the root (averaged over every random tree) are more likely to be anomalies. 
\item {\bf The Local Outlier Factor (LOF)}~\cite{breunig_lof_2000} computes the ratio of the neighbor density to the local density. 
\item {\bf Matrix Profile (MP)}~\cite{yeh_time_2018} detects as anomaly the subsequence with the most significant 1-NN distance. 
\item {\bf NormA}~\cite{boniol_unsupervised_2021} identifies the normal patterns based on clustering and calculates each point's distance to normal patterns weighted using statistical criteria. 
\item {\bf Principal Component Analysis (PCA)}~\cite{aggarwal_outlier_2017} projects data to a lower-dimensional hyperplane. Outliers are points with a large distance from this plane. 
\item {\bf Autoencoder (AE)} \cite{10.1145/2689746.2689747} projects data to a lower-dimensional space and reconstructs it. Outliers are expected to have larger reconstruction errors. 
\item {\bf LSTM-AD}~\cite{malhotra_long_2015} use an LSTM network that predicts future values from the current subsequence. The prediction error is used to identify anomalies.
\item {\bf Polynomial Approximation (POLY)} \cite{li_unifying_2007} fits a polynomial model that tries to predict the values of the data series from the previous subsequences. Outliers are detected with the prediction error. 
\item {\bf CNN} \cite{8581424} built, using a convolutional deep neural network, a correlation between current and previous subsequences, and outliers are detected by the deviation between the prediction and the actual value. 
\item {\bf One-class Support Vector Machines (OCSVM)} \cite{scholkopf_support_1999} is a support vector method that fits a training dataset and finds the normal data's boundary.
\end{itemize}

\subsection{Qualitative Analysis}
\label{exp:qual}

We first use two examples to demonstrate qualitatively the limitations of existing accuracy evaluation measures in the presence of lag and noise, and to motivate the need for a new approach. 
These two examples are depicted in Figure~\ref{fig:quality}. 
The first example, in Figure~\ref{fig:quality}(A), corresponds to OCSVM and AE on the MBA(805) dataset (named MBA\_ECG805\_data.out in the ECG dataset). 

We observe in Figure~\ref{fig:quality}(A)(a.1) and (a.2) that both scores identify most of the anomalies (highlighted in red). However, the OCSVM score points to more false positives (at the end of the time series) and only captures small sections of the anomalies. On the contrary, the AE score points to fewer false positives and captures all abnormal subsequences. Thus we can conclude that, visually, AE should obtain a better accuracy score than OCSVM. Nevertheless, we also observe that the AE score is lagged with the labels and contains more noise. The latter has a significant impact on the accuracy of evaluation measures. First, Figure~\ref{fig:quality}(A)(c) is showing that AUC-PR is better for OCSM (0.73) than for AE (0.57). This is contradictory with what is visually observed from Figure~\ref{fig:quality}(A)(a.1) and (a.2). However, when using our proposed measure R-AUC-PR, OCSVM obtains a lower score (0.83) than AE (0.89). This confirms that, in this example, a buffer region before the labels helps to capture the true value of an anomaly score. Overall, Figure~\ref{fig:quality}(A)(f) is showing in green and red the evolution of accuracy score for the 13 accuracy measures for AE and OCSVM, respectively. The latter shows that, in addition to Precision@k and Precision, our proposed approach captures the quality order between the two methods well.

We now present a second example, on a different time series, illustrated in Figure~\ref{fig:quality}(B). 
In this case, we demonstrate the anomaly score of OCSVM and LOF (depicted in Figure~\ref{fig:quality}(B)(a.1) and (a.2)) applied on the MBA(806) dataset (named MBA\_ECG806\_data.out in the ECG dataset). 
We observe that both methods produce the same level of noise. However, LOF points to fewer false positives and captures more sections of the abnormal subsequences than OCSVM. 
Nevertheless, the LOF score is slightly lagged with the labels such that the maximum values in the LOF score are slightly outside of the labeled sections. 
Thus, as illustrated in Figure~\ref{fig:quality}(B)(f), even though we can visually consider that LOF is performing better than OCSM, all usual measures (Precision, Recall, F, precision@k, and AUC-PR) are judging OCSM better than AE. On the contrary, measures that consider lag (Rprecision, Rrecall, RF) rank the methods correctly. 
However, due to threshold issues, these measures are very close for the two methods. Overall, only AUC-ROC and our proposed measures give a higher score for LOF than for OCSVM.

\subsection{Quantitative Analysis}
\label{exp:case}

\begin{figure}[t]
  \centering
  \includegraphics[width=1\linewidth]{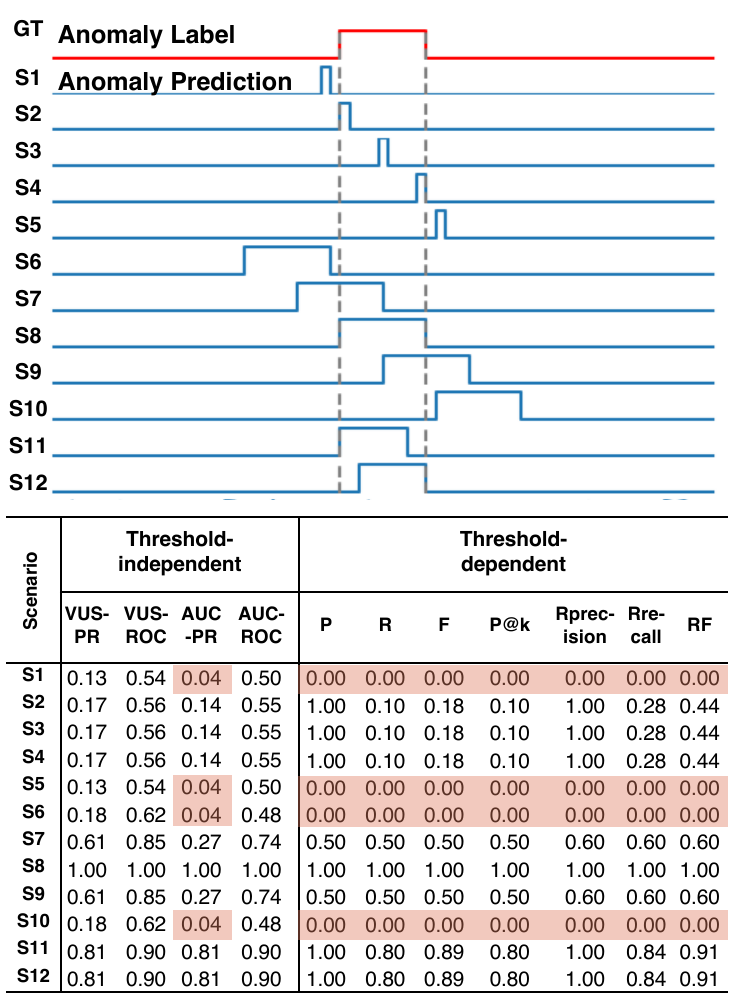}
  %\vspace*{-0.7cm}
  \caption{\commentRed{
  Comparison of evaluation measures for synthetic data examples across various scenarios. S8 represents the oracle case, where predictions perfectly align with labeled anomalies. Problematic cases are highlighted in the red region.}}
  %\vspace*{-0.5cm}
  \label{fig:eval_case_study}
\end{figure}
\commentRed{
We present the evaluation results for different synthetic data scenarios, as shown in Figure~\ref{fig:eval_case_study}. These scenarios range from S1, where predictions occur before the ground truth anomaly, to S12, where predictions fall within the ground truth region. The red-shaded regions highlight problematic cases caused by a lack of adaptability to lags. For instance, in scenarios S1 and S2, a slight shift in the prediction leads to measures (e.g., AUC-PR, F score) that fail to account for lags, resulting in a zero score for S1 and a significant discrepancy between the results of S1 and S2. Thus, we observe that our proposed VUS effectively addresses these issues and provides robust evaluations results.}

%\subsection{Quantitative Analysis}
%\subsection{Sensitivity and Separability Analysis}
\subsection{Robustness Analysis}
\label{exp:quant}

\begin{figure}[tb]
  \centering
  \includegraphics[width=1\linewidth]{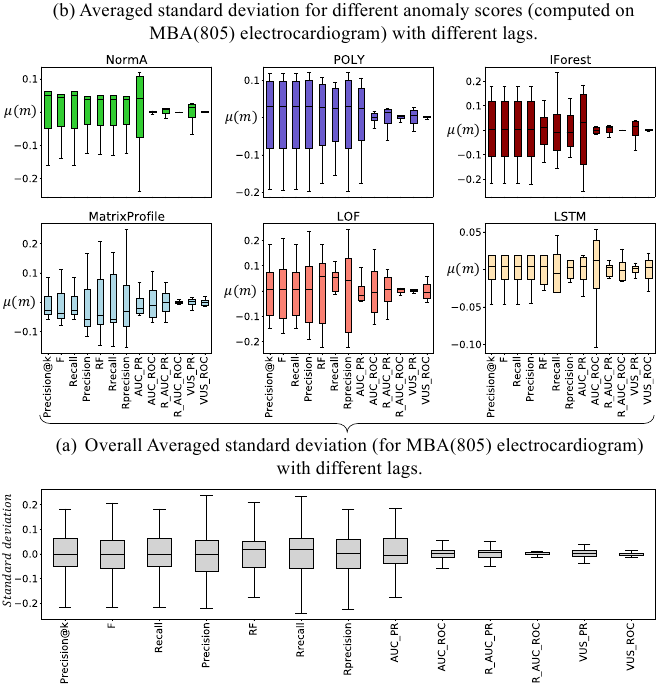}
  %\vspace*{-0.7cm}
  \caption{For each method, we compute the accuracy measures 10 times with random lag $\ell \in [-0.25*\ell,0.25*\ell]$ injected in the anomaly score. We center the accuracy average to 0.}
  %\vspace*{-0.5cm}
  \label{fig:lagsensitivity}
\end{figure}

\begin{figure}[tb]
  \centering
  \includegraphics[width=\linewidth]{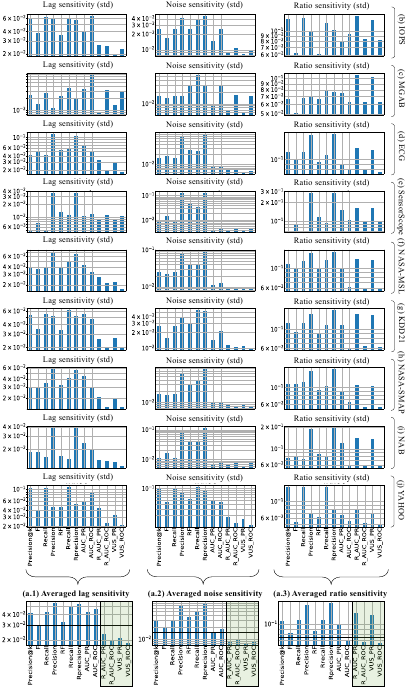}
%  %\vspace*{-0.3cm}
  \caption{Robustness Analysis for nine datasets: we report, over the entire benchmark, the average standard deviation of the accuracy values of the measures, under varying (a.1) lag, (a.2) noise, and (a.3) normal/abnormal ratio. }
  \label{fig:sensitivity_per_data}
\end{figure}

\begin{figure*}[tb]
  \centering
  \includegraphics[width=\linewidth]{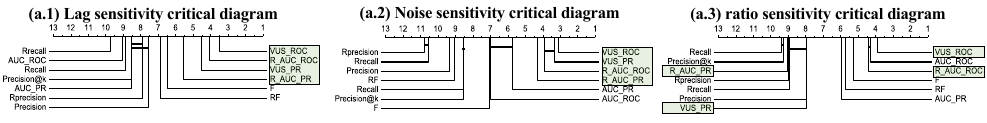}
  %\vspace*{-0.7cm}
  \caption{Critical difference diagram computed using the signed-rank Wilkoxon test (with $\alpha=0.1$) for the robustness to (a.1) lag, (a.2) noise and (a.3) normal/abnormal ratio.}
  \label{fig:sensitivity}
\end{figure*}

We have illustrated with specific examples several of the limitations of current measures. 
We now evaluate quantitatively the robustness of the proposed measures when compared to the currently used measures. 
We first evaluate the robustness to noise, lag, and normal versus abnormal points ratio. We then measure their ability to separate accurate and inaccurate methods.
%\newline \textbf{Sensitivity Analysis: } 
We first analyze the robustness of different approaches quantitatively to different factors: (i) lag, (ii) noise, and (iii) normal/abnormal ratio. As already mentioned, these factors are realistic. For instance, lag can be either introduced by the anomaly detection methods (such as methods that produce a score per subsequences are only high at the beginning of abnormal subsequences) or by human labeling approximation. Furthermore, even though lag and noises are injected, an optimal evaluation metric should not vary significantly. Therefore, we aim to measure the variance of the evaluation measures when we vary the lag, noise, and normal/abnormal ratio. We proceed as follows:

\begin{enumerate}[noitemsep,topsep=0pt,parsep=0pt,partopsep=0pt,leftmargin=0.5cm]
\item For each anomaly detection method, we first compute the anomaly score on a given time series.
\item We then inject either lag $l$, noise $n$ or change the normal/abnormal ratio $r$. For 10 different values of $l \in [-0.25*\ell,0.25*\ell]$, $n \in [-0.05*(max(S_T)-min(S_T)),0.05*(max(S_T)-min(S_T))]$ and $r \in [0.01,0.2]$, we compute the 13 different measures.
\item For each evaluation measure, we compute the standard deviation of the ten different values. Figure~\ref{fig:lagsensitivity}(b) depicts the different lag values for six AD methods applied on a data series in the ECG dataset.
\item We compute the average standard deviation for the 13 different AD quality measures. For example, figure~\ref{fig:lagsensitivity}(a) depicts the average standard deviation for ten different lag values over the AD methods applied on the MBA(805) time series.
\item We compute the average standard deviation for the every time series in each dataset (as illustrated in Figure~\ref{fig:sensitivity_per_data}(b to j) for nine datasets of the benchmark.
\item We compute the average standard deviation for the every dataset (as illustrated in Figure~\ref{fig:sensitivity_per_data}(a.1) for lag, Figure~\ref{fig:sensitivity_per_data}(a.2) for noise and Figure~\ref{fig:sensitivity_per_data}(a.3) for normal/abnormal ratio).
\item We finally compute the Wilcoxon test~\cite{10.2307/3001968} and display the critical diagram over the average standard deviation for every time series (as illustrated in Figure~\ref{fig:sensitivity}(a.1) for lag, Figure~\ref{fig:sensitivity}(a.2) for noise and Figure~\ref{fig:sensitivity}(a.3) for normal/abnormal ratio).
\end{enumerate}

The methods with the smallest standard deviation can be considered more robust to lag, noise, or normal/abnormal ratio from the above framework. 
First, as stated in the introduction, we observe that non-threshold-based measures (such as AUC-ROC and AUC-PR) are indeed robust to noise (see Figure~\ref{fig:sensitivity_per_data}(a.2)), but not to lag. Figure~\ref{fig:sensitivity}(a.1) demonstrates that our proposed measures VUS-ROC, VUS-PR, R-AUC-ROC, and R-AUC-PR are significantly more robust to lag. Similarly, Figure~\ref{fig:sensitivity}(a.2) confirms that our proposed measures are significantly more robust to noise. However, we observe that, among our proposed measures, only VUS-ROC and R-AUC-ROC are robust to the normal/abnormal ratio and not VUS-PR and R-AUC-PR. This is explained by the fact that Precision-based measures vary significantly when this ratio changes. This is confirmed by Figure~\ref{fig:sensitivity_per_data}(a.3), in which we observe that Precision and Rprecision have a high standard deviation. Overall, we observe that VUS-ROC is significantly more robust to lag, noise, and normal/abnormal ratio than other measures.

\subsection{Separability Analysis}
\label{exp:separability}

\begin{figure*}[tb]
  \centering
  \includegraphics[width=1\linewidth]{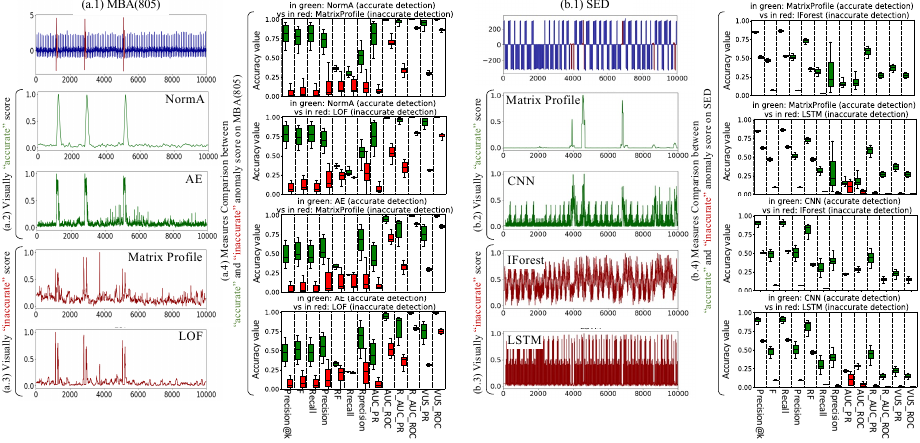}
  %\vspace*{-0.5cm}
  \caption{Separability analysis applied on 4 pairs of accurate (green) and inaccurate (red) methods on (a) the MBA(805) data series, and (b) the SED data series.}
  %\vspace*{-0.3cm}
  \label{fig:separability}
\end{figure*}

%\newline \textbf{Separability Analysis: } 
We now evaluate the separability capacities of the different evaluation metrics. 
\commentRed{The main objective is to measure the ability of accuracy measures to separate accurate methods from inaccurate ones. More precisely, an appropriate measure should return accuracy scores that are significantly higher for accurate anomaly scores than for inaccurate ones.}
We thus manually select accurate and inaccurate anomaly detection methods and verify if the accuracy evaluation scores are indeed higher for the accurate than for the inaccurate methods. Figure~\ref{fig:separability} depicts the latter separability analysis applied to the MBA(805) and the SED series. 
The accurate and inaccurate anomaly scores are plotted in green and red, respectively. 
We then consider 12 different pairs of accurate/inaccurate methods among the eight previously mentioned anomaly scores. 
We slightly modify each score 50 different times in which we inject lag and noises and compute the accuracy measures. 
Figure~\ref{fig:separability}(a.4) and Figure~\ref{fig:separability}(b.4) are divided into four different subplots corresponding to 4 pairs (selected among the twelve different pairs due to lack of space). 
Each subplot corresponds to two box plots per accuracy measure. 
The green and red box plots correspond to the 50 accuracy measures on the accurate and inaccurate methods. 
If the red and green box plots are well separated, we can conclude that the corresponding accuracy measures are separating the accurate and inaccurate methods well. 
We observe that some accuracy measures (such as VUS-ROC) are more separable than others (such as RF). We thus measure the separability of the two box-plots by computing the Z-test. 

\begin{figure}[tb]
  \centering
  \includegraphics[width=\linewidth]{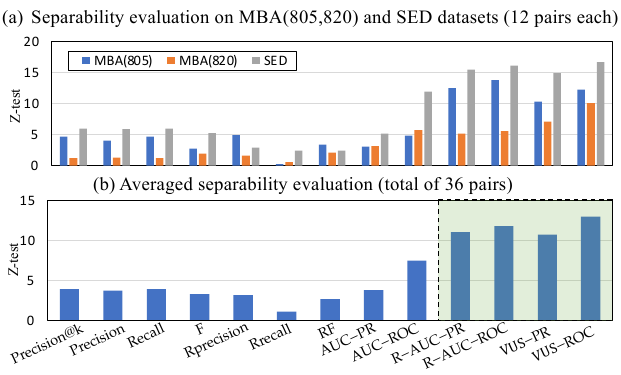}
  %\vspace*{-0.5cm}
  \caption{Overall separability analysis (averaged z-test between the accuracy values distributions of accurate and inaccurate methods) applied on 36 pairs on 3 datasets.}
  \label{fig:separability_agg}
\end{figure}

We now aggregate all the results and compute the average Z-test for all pairs of accurate/inaccurate datasets (examples are shown in Figures~\ref{fig:separability}(a.2) and (b.2) for accurate anomaly scores, and in Figures~\ref{fig:separability}(a.3) and (b.3) for inaccurate anomaly scores, for the MBA(805) and SED series, respectively). 
Next, we perform the same operation over three different data series: MBA (805), MBA(820), and SED. 
Then, we depict the average Z-test for these three datasets in Figure~\ref{fig:separability_agg}(a). 
Finally, we show the average Z-test for all datasets in Figure~\ref{fig:separability_agg}(b).

We observe that our proposed VUS-based and Range-based measures are significantly more separable than other current accuracy measures (up to two times for AUC-ROC, the best measures of all current ones). Furthermore, when analyzed in detail in Figure~\ref{fig:separability} and Figure~\ref{fig:separability_agg}, we confirm that VUS-based and Range-based are more separable over all three datasets. 

\noindent \textbf{Global Analysis: } Overall, we observe that VUS-ROC is the most robust (cf. Figure~\ref{fig:sensitivity}) and separable (cf. Figure~\ref{fig:separability_agg}) measure. 
On the contrary, Precision and Rprecision are non-robust and non-separable. 
Among all previous accuracy measures, only AUC-ROC is robust and separable. 
Popular measures, such as, F, RF, AUC-ROC, and AUC-PR are robust but non-separable.

\begin{figure}[tb]
  \centering
  \includegraphics[width=\linewidth]{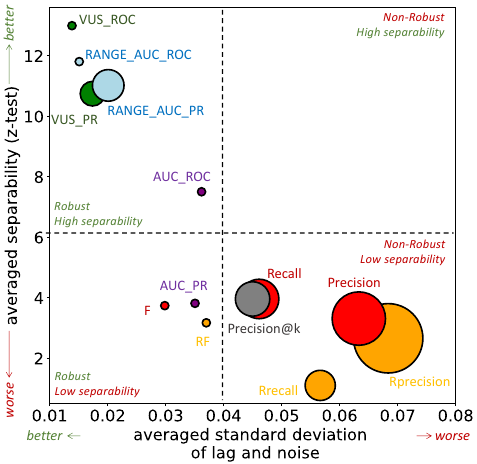}
  \caption{Evaluation of all measures based on: (y-axis) their separability (avg. z-test), (x-axis) avg. standard deviation of the accuracy values when varying lag and noise, (circle size) avg. standard deviation of the accuracy values when varying the normal/abnormal ratio.}
  \label{fig:global}
\end{figure}

In order to visualize the global statistical analysis, we merge the robustness and the separability analysis into a single plot. Figure~\ref{fig:global} depicts one scatter point per accuracy measure. 
The x-axis represents the averaged standard deviation of lag and noise (averaged values from Figure~\ref{fig:sensitivity_per_data}(a.1) and (a.2)). The y-axis corresponds to the averaged Z-test (averaged value from Figure~\ref{fig:separability_agg}). 
Finally, the size of the points corresponds to the sensitivity to the normal/abnormal ratio (values from Figure~\ref{fig:sensitivity_per_data}(a.3)). 
Figure~\ref{fig:global} demonstrates that our proposed measures (located at the top left section of the plot) are both the most robust and the most separable. 
Among all previous accuracy measures, only AUC-ROC is on the top left section of the plot. 
Popular measures, such as, F, RF, AUC-ROC, AUC-PR are on the bottom left section of the plot. 
The latter underlines the fact that these measures are robust but non-separable.
Overall, Figure~\ref{fig:global} confirms the effectiveness and superiority of our proposed measures, especially of VUS-ROC and VUS-PR.

\subsection{Consistency Analysis}
\label{sec:entropy}

In this section, we analyze the accuracy of the anomaly detection methods provided by the 13 accuracy measures. The objective is to observe the changes in the global ranking of anomaly detection methods. For that purpose, we formulate the following assumptions. First, we assume that the data series in each benchmark dataset are similar (i.e., from the same domain and sharing some common characteristics). As a matter of fact, we can assume that an anomaly detection method should perform similarly on these data series of a given dataset. This is confirmed when observing that the best anomaly detection methods are not the same based on which dataset was analyzed. Thus the ranking of the anomaly detection methods should be different for different datasets, but similar for every data series in each dataset. 
Therefore, for a given method $A$ and a given dataset $D$ containing data series of the same type and domain, we assume that a good accuracy measure results in a consistent rank for the method $A$ across the dataset $D$. 
The consistency of a method's ranks over a dataset can be measured by computing the entropy of these ranks. 
For instance, a measure that returns a random score (and thus, a random rank for a method $A$) will result in a high entropy. 
On the contrary, a measure that always returns (approximately) the same ranks for a given method $A$ will result in a low entropy. 
Thus, for a given method $A$ and a given dataset $D$ containing data series of the same type and domain, we assume that a good accuracy measure results in a low entropy for the different ranks for method $A$ on dataset $D$.

\begin{figure*}[tb]
  \centering
  \includegraphics[width=\linewidth]{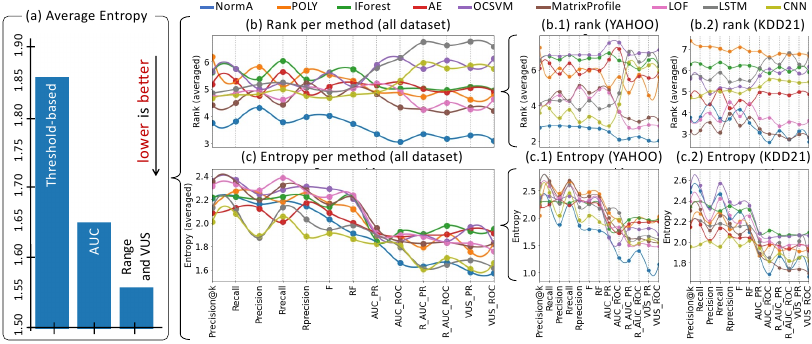}
  %\vspace*{-0.5cm}
  \caption{Accuracy evaluation of the anomaly detection methods. (a) Overall average entropy per category of measures. Analysis of the (b) averaged rank and (c) averaged rank entropy for each method and each accuracy measure over the entire benchmark. Example of (b.1) average rank and (c.1) entropy on the YAHOO dataset, KDD21 dataset (b.2, c.2). }
  \label{fig:entropy}
\end{figure*}

We now compute the accuracy measures for the nine different methods (we compute the anomaly scores ten different times, and we use the average accuracy). 
Figures~\ref{fig:entropy}(b.1) and (b.2) report the average ranking of the anomaly detection methods obtained on the YAHOO and KDD21 datasets, respectively. 
The x-axis corresponds to the different accuracy measures. We first observe that the rankings are more separated using Range-AUC and VUS measures for these two datasets. Figure~\ref{fig:entropy}(b) depicts the average ranking over the entire benchmark. The latter confirms the previous observation that VUS measures provide more separated rankings than threshold-based and AUC-based measures. We also observe an interesting ranking evolution for the YAHOO dataset illustrated in Figure~\ref{fig:entropy}(b.1). We notice that both LOF and MatrixProfile (brown and pink curve) have a low rank (between 4 and 5) using threshold and AUC-based measures. However, we observe that their ranks increase significantly for range-based and VUS-based measures (between 2.5 and 3). As we noticed by looking at specific examples (see Figure~\ref{exp:qual}), LOF and MatrixProfile can suffer from a lag issue even though the anomalies are well-identified. Therefore, the range-based and VUS-based measures better evaluate these two methods' detection capability.

Overall, the ranking curves show that the ranks appear more chaotic for threshold-based than AUC-, Range-AUC-, and VUS-based measures. 
In order to quantify this observation, we compute the Shannon Entropy of the ranks of each anomaly detection method. 
In practice, we extract the ranks of methods across one dataset and compute Shannon's Entropy of the different ranks. 
Figures~\ref{fig:entropy}(c.1) and (c.2) depict the entropy of each of the nine methods for the YAHOO and KDD21 datasets, respectively. 
Figure~\ref{fig:entropy}(c) illustrates the averaged entropy for all datasets in the benchmark for each measure and method, while Figure~\ref{fig:entropy}(a) shows the averaged entropy for each category of measures.
We observe that both for the general case (Figure~\ref{fig:entropy}(a) and Figure~\ref{fig:entropy}(c)) and some specific cases (Figures~\ref{fig:entropy}(c.1) and (c.2)), the entropy is reducing when using AUC-, Range-AUC-, and VUS-based measures. 
We report the lowest entropy for VUS-based measures. 
Moreover, we notice a significant drop between threshold-based and AUC-based. 
This confirms that the ranks provided by AUC- and VUS-based measures are consistent for data series belonging to one specific dataset.

Therefore, based on the assumption formulated at the beginning of the section, we can thus conclude that AUC, range-AUC, and VUS-based measures are providing more consistent rankings. Finally, as illustrated in Figure~\ref{fig:entropy}, we also observe that VUS-based measures result in the most ordered and similar rankings for data series from the same type and domain.

\subsection{Execution Time Analysis}
\label{sec:exectime}

In this section, we evaluate the execution time required to compute different evaluation measures. 
In Section~\ref{sec:synthetic_eval_time}, we first measure the influence of different time series characteristics and VUS parameters on the execution time. In Section~\ref{sec:TSB_eval_time}, we  measure the execution time of VUS (VUS-ROC and VUS-PR simultaneously), R-AUC (R-AUC-ROC and R-AUC-PR simultaneously), and AUC-based measures (AUC-ROC and AUC-PR simultaneously) on the TSB-UAD benchmark. \commentRed{As demonstrated in the previous section, threshold-based measures are not robust, have a low separability power, and are inconsistent. 
Such measures are not suitable for evaluating anomaly detection methods. Thus, in this section, we do not consider threshold-based measures.}

\subsubsection{Evaluation on Synthetic Time Series}\hfill\\
\label{sec:synthetic_eval_time}

We first analyze the impact that time series characteristics and parameters have on the computation time of VUS-based measures. 
to that effect, we generate synthetic time series and labels, where we vary the following parameters: (i) the number of anomalies {\bf$\alpha$} in the time series, (ii) the average \textbf{$\mu(\ell_a)$} and standard deviation $\sigma(\ell_a)$ of the anomalies lengths in the time series (all the anomalies can have different lengths), (iii) the length of the time series \textbf{$|T|$}, (iv) the maximum buffer length \textbf{$L$}, and (v) the number of thresholds \textbf{$N$}.

We also measure the influence on the execution time of the R-AUC- and AUC- related parameter, that is, the number of thresholds ($N$).
The default values and the range of variation of these parameters are listed in Table~\ref{tab:parameter_range_time}. 
For VUS-based measures, we evaluate the execution time of the initial VUS implementation, as well as the two optimized versions, VUS$_{opt}$ and VUS$_{opt}^{mem}$.

\begin{table}[tb]
    \centering
    \caption{Value ranges for the parameters: number of anomalies ($\alpha$), average and standard deviation anomaly length ($\mu(\ell_a)$,$\sigma(\ell_a)$), time series length ($|T|$), maximum buffer length ($L$), and number of thresholds ($N$).}
    \begin{tabular}{|c|c|c|c|c|c|c|} 
 \hline
 Param. & $\alpha$ & $\mu(\ell_a)$ & $\sigma(\ell_{a})$ & $|T|$ & $L$ & $N$ \\ [0.5ex] 
 \hline\hline
 \textbf{Default} & 10 & 10 & 0 & $10^5$ & 5 & 250\\ 
 \hline
 Min. & 0 & 0 & 0 & $10^3$ & 0 & 2 \\
 \hline
 Max. & $2*10^3$ & $10^3$ & $10$ & $10^5$ & $10^3$ & $10^3$ \\ [1ex] 
 \hline
\end{tabular}
    \label{tab:parameter_range_time}
\end{table}

\begin{figure*}[tb]
  \centering
  \includegraphics[width=\linewidth]{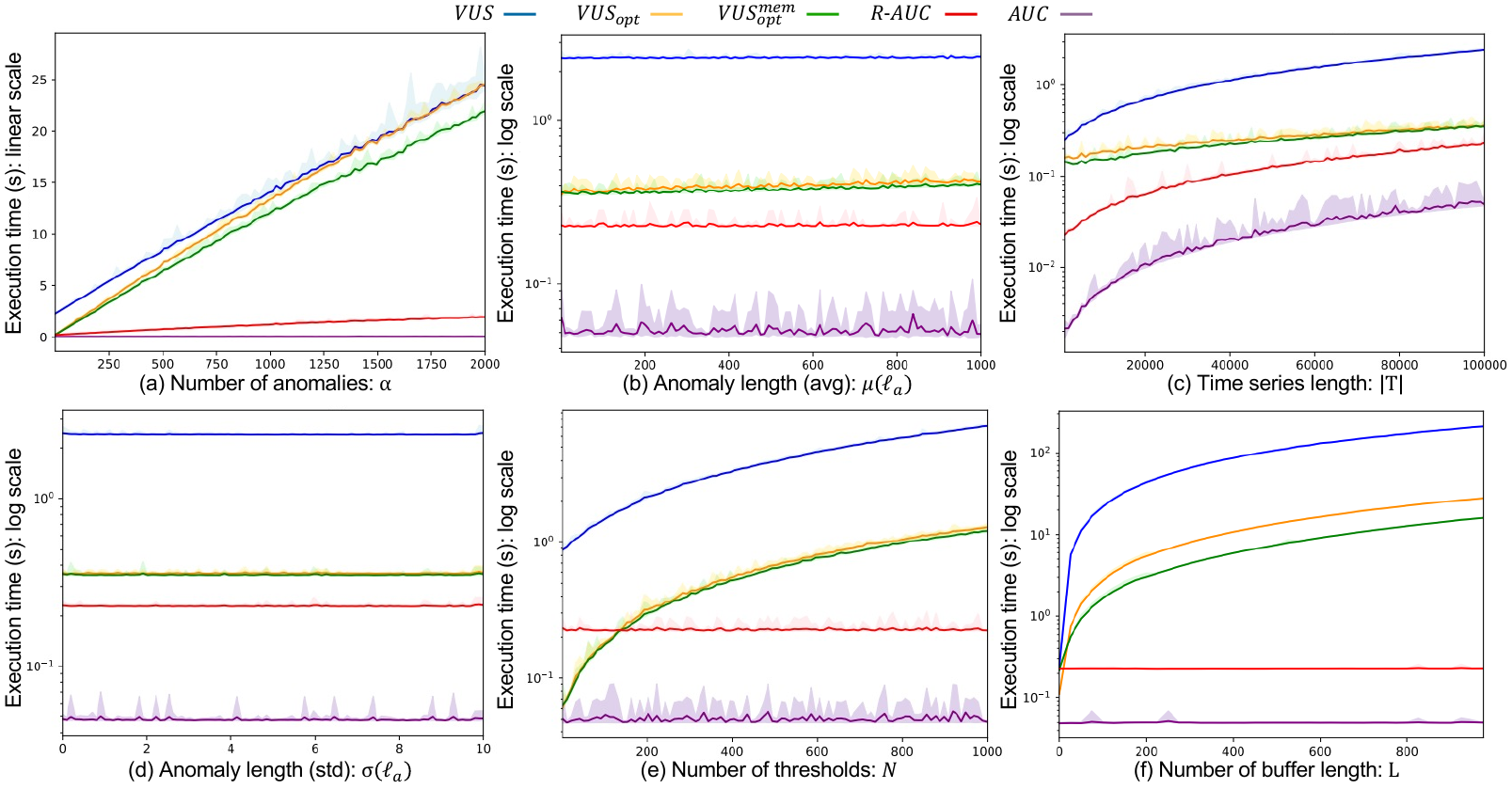}
  %\vspace*{-0.5cm}
  \caption{Execution time of VUS, R-AUC, AUC-based measures when we vary the parameters listed in Table~\ref{tab:parameter_range_time}. The solid lines correspond to the average execution time over 10 runs. The colored envelopes are to the standard deviation.}
  \label{fig:sythetic_exp_time}
\end{figure*}

Figure~\ref{fig:sythetic_exp_time} depicts the execution time (averaged over ten runs) for each parameter listed in Table~\ref{tab:parameter_range_time}. 
Overall, we observe that the execution time of AUC-based and R-AUC-based measures is significantly smaller than VUS-based measures.
In the following paragraph, we analyze the influence of each parameter and compare the experimental execution time evaluation to the theoretical complexity reported in Table~\ref{tab:complexity_summary}.

\vspace{0.2cm}
\noindent {\bf [Influence of $\alpha$]}:
In Figure~\ref{fig:sythetic_exp_time}(a), we observe that the VUS, VUS$_{opt}$, and VUS$_{opt}^{mem}$ execution times are linearly increasing with $\alpha$. 
The increase in execution time for VUS, VUS$_{opt}$, and VUS$_{opt}^{mem}$ is more pronounced when we vary $\alpha$, in contrast to $l_a$ (which nevertheless, has a similar effect on the overall complexity). 
We also observe that the VUS$_{opt}^{mem}$ execution time grows slower than $VUS_{opt}$ when $\alpha$ increases. 
This is explained by the use of 2-dimensional arrays for the storage of predictions, which use contiguous memory locations that allow for faster access, decreasing the dependency on $\alpha$.

\vspace{0.2cm}
\noindent {\bf [Influence of $\mu(\ell_a)$]}:
As shown in Figure~\ref{fig:sythetic_exp_time}(b), the execution time variation of VUS, VUS$_{opt}$, and VUS$_{opt}^{mem}$ caused by $\ell_a$ is rather insignificant. 
We also observe that the VUS$_{opt}$ and VUS$_{opt}^{mem}$ execution times are significantly lower when compared to VUS. 
This is explained by the smaller dependency of the complexity of these algorithms on the time series length $|T|$. 
Overall, the execution time for both VUS$_{opt}$ and VUS$_{opt}^{mem}$ is significantly lower than VUS, and follows a similar trend. 

\vspace{0.2cm}
\noindent {\bf [Influence of $\sigma(\ell_a)$]}: 
As depicted in Figure~\ref{fig:sythetic_exp_time}(d) and inferred from the theoretical complexities in Table~\ref{tab:complexity_summary}, none of the measures are affected by the standard deviation of the anomaly lengths.

\vspace{0.2cm}
\noindent {\bf [Influence of $|T|$]}:
For short time series (small values of $|T|$), we note that O($T_1$) becomes comparable to O($T_2$). 
Thus, the theoretical complexities approximate to $O(NL(T_1+T_2))$, $O(N*(T_1+T_2))+O(NLT_2)$ and $O(N(T_1+T_2))$ for VUS, VUS$_{opt}$, and VUS$_{opt}^{mem}$, respectively. 
Indeed, we observe in Figure~\ref{fig:sythetic_exp_time}(c) that the execution times of VUS, VUS$_{opt}$, and VUS$_{opt}^{mem}$ are similar for small values of $|T|$. However, for larger values of $|T|$, $O(T_1)$ is much higher compared to $O(T_2)$, thus resulting in an effective complexity of $O(NLT_1)$ for VUS, and $O(NT_1)$ for VUS$_{opt}$, and VUS$_{opt}^{mem}$. 
This translates to a significant improvement in execution time complexity for VUS$_{opt}$ and VUS$_{opt}^{mem}$ compared to VUS, which is confirmed by the results in Figure~\ref{fig:sythetic_exp_time}(c).

\vspace{0.2cm}
\noindent {\bf [Influence of $N$]}: 
Given the theoretical complexity depicted in Table~\ref{tab:complexity_summary}, it is evident that the number of thresholds affects all measures in a linear fashion.
Figure~\ref{fig:sythetic_exp_time}(e) demonstrates this point: the results of varying $N$ show a linear dependency for VUS, VUS$_{opt}$, and VUS$_{opt}^{mem}$ (i.e., a logarithmic trend with a log scale on the y axis). \commentRed{Moreover, we observe that the AUC and range-AUC execution time is almost constant regardless of the number of thresholds used. The latter is explained by the very efficient implementation of AUC measures. Therefore, the linear dependency on the number of thresholds is not visible in Figure~\ref{fig:sythetic_exp_time}(e).}

\vspace{0.2cm}
\noindent {\bf [Influence of $L$]}: Figure~\ref{fig:sythetic_exp_time}(f) depicts the influence of the maximum buffer length $L$ on the execution time of all measures. 
We observe that, as $L$ grows, the execution time of VUS$_{opt}$ and VUS$_{opt}^{mem}$ increases slower than VUS. 
We also observe that VUS$_{opt}^{mem}$ is more scalable with $L$ when compared to VUS$_{opt}$. 
This is consistent with the theoretical complexity (cf. Table~\ref{tab:complexity_summary}), which indicates that the dependence on $L$ decreases from $O(NL(T_1+T_2+\ell_a \alpha))$ for VUS to $O(NL(T_2+\ell_a \alpha)$ and $O(NL(\ell_a \alpha))$ for $VUS_{opt}$, and $VUS_{opt}^{mem}$.

\vspace{0.2cm}
In order to obtain a more accurate picture of the influence of each of the above parameters, we fit the execution time (as affected by the parameter values) using linear regression; we can then use the regression slope coefficient of each parameter to evaluate the influence of that parameter. 
In practice, we fit each parameter individually, and report the regression slope coefficient, as well as the coefficient of determination $R^2$.
Table~\ref{tab:parameter_linear_coeff} reports the coefficients mentioned above for each parameter associated with VUS, VUS$_{opt}$, and VUS$_{opt}^{mem}$.

\begin{table}[tb]
    \centering
    \caption{Linear regression slope coefficients ($C.$) for VUS execution times, for each parameter independently. }
    \begin{tabular}{|c|c|c|c|c|c|c|} 
 \hline
 Measure & Param. & $\alpha$ & $l_a$ & $|T|$ & $L$ & $N$\\ [0.5ex] 
 \hline\hline
 \multirow{2}{*}{$VUS$} & $C.$ & 21.9 & 0.02 & 2.13 & 212 & 6.24\\\cline{2-7}
 & {$R^2$} & 0.99 & 0.15 & 0.99 & 0.99 & 0.99 \\   
 \hline
  \multirow{2}{*}{$VUS_{opt}$} & $C.$ & 24.2  & 0.06 & 0.19 & 27.8 & 1.23\\\cline{2-7}
  & $R^2$& 0.99 & 0.86 & 0.99 & 0.99 & 0.99\\ 
 \hline
 \multirow{2}{*}{$VUS_{opt}^{mem}$} & $C.$ & 21.5 & 0.05 & 0.21 & 15.7 & 1.16\\\cline{2-7}
  & $R^2$ & 0.99 & 0.89 & 0.99 & 0.99 & 0.99\\[1ex] 
 \hline
\end{tabular}
    \label{tab:parameter_linear_coeff}
\end{table}

Table~\ref{tab:parameter_linear_coeff} shows that the linear regression between $\alpha$ and the execution time has a $R^2=0.99$. Thus, the dependence of execution time on $\alpha$ is linear. We also observe that VUS$_{opt}$ execution time is more dependent on $\alpha$ than VUS and VUS$_{opt}^{mem}$ execution time.
Moreover, the dependence of the execution time on the time series length ($|T|$) is higher for VUS than for VUS$_{opt}$ and VUS$_{opt}^{mem}$. 
More importantly, VUS$_{opt}$ and VUS$_{opt}^{mem}$ are significantly less dependent than VUS on the number of thresholds and the maximal buffer length.

\subsubsection{Evaluation on TSB-UAD Time Series}\hfill\\
\label{sec:TSB_eval_time}

In this section, we verify the conclusions outlined in the previous section with real-world time series from the TSB-UAD benchmark. 
In this setting, the parameters $\alpha$, $\ell_a$, and $|T|$ are calculated from the series in the benchmark and cannot be changed. Moreover, $L$ and $N$ are parameters for the computation of VUS, regardless of the time series (synthetic or real). Thus, we do not consider these two parameters in this section.

\begin{figure*}[tb]
  \centering
  \includegraphics[width=\linewidth]{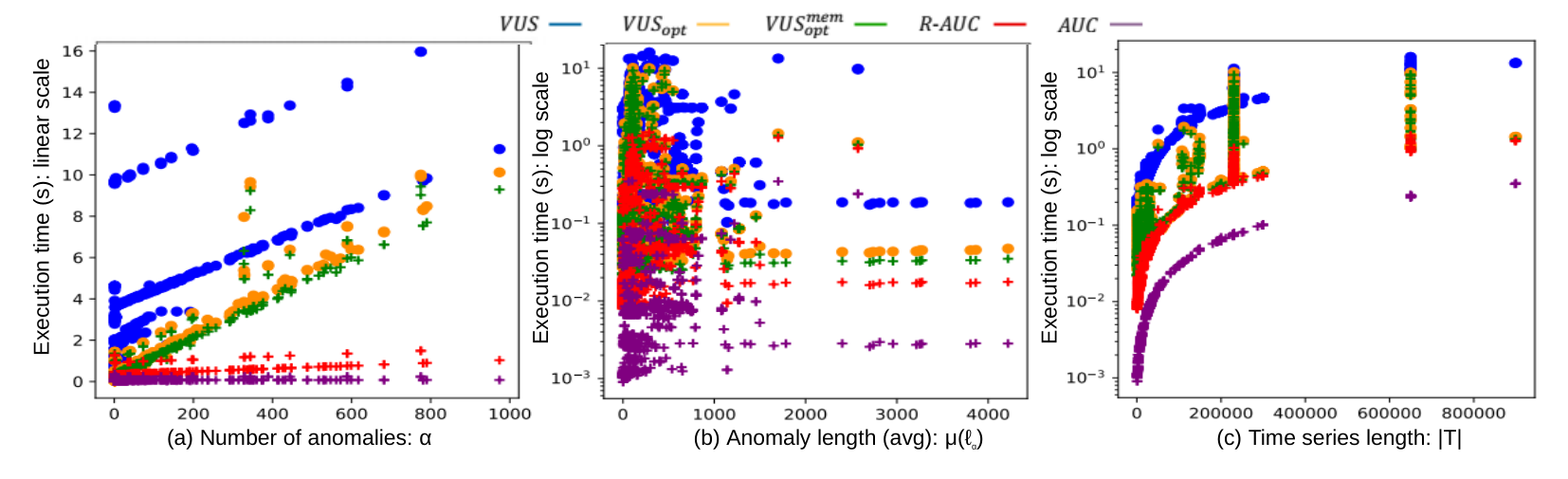}
  \caption{Execution time of VUS, R-AUC, AUC-based measures on the TSB-UAD benchmark, versus $\alpha$, $\ell_a$, and $|T|$.}
  \label{fig:TSB}
\end{figure*}

Figure~\ref{fig:TSB} depicts the execution time of AUC, R-AUC, and VUS-based measures versus $\alpha$, $\mu(\ell_a)$, and $|T|$.
We first confirm with Figure~\ref{fig:TSB}(a) the linear relationship between $\alpha$ and the execution time for VUS, VUS$_{opt}$ and VUS$_{opt}^{mem}$.
On further inspection, it is possible to see two separate lines for almost all the measures. 
These lines can be attributed to the time series length $|T|$. 
The convergence of VUS and $VUS_{opt}$ when $\alpha$ grows shows the stronger dependence that $VUS_{opt}$ execution time has on $\alpha$, as already observed with the synthetic data (cf. Section~\ref{sec:synthetic_eval_time}). 

In Figure~\ref{fig:TSB}(b), we observe that the variation of the execution time with $\ell_a$ is limited when compared to the two other parameters. We conclude that the variation of $\ell_a$ is not a key factor in determining the execution time of the measures.
Furthermore, as depicted in Figure~\ref{fig:TSB}(c), $VUS_{opt}$ and $VUS_{opt}^{mem}$ are more scalable than VUS when $|T|$ increases. 
We also confirm the linear dependence of execution time on the time series length for all the accuracy measures, which is consistent with the experiments on the synthetic data. 
The two abrupt jumps visible in Figure~\ref{fig:TSB}(c) are explained by significant increases of $\alpha$ in time series of the same length. 

\begin{table}[tb]
\centering
\caption{Linear regression slope coefficients ($C.$) for VUS execution time, for all time series parameters all-together.}
\begin{tabular}{|c|ccc|c|} 
 \hline
Measure & $\alpha$ & $|T|$ & $l_a$ & $R^2$ \\ [0.5ex] 
 \hline\hline
 \multirow{1}{*}{${VUS}$} & 7.87 & 13.5 & -0.08 & 0.99  \\ 
 %\cline{2-5} & $R^2$ & \multicolumn{3}{c|}{ 0.99}\\
 \hline
 \multirow{1}{*}{$VUS_{opt}$} & 10.2 & 1.70 & 0.09 & 0.96 \\
 %\cline{2-5} & $R^2$ & \multicolumn{3}{c|}{0.96}\\
\hline
 \multirow{1}{*}{$VUS_{opt}^{mem}$} & 9.27 & 1.60 & 0.11 & 0.96 \\
 %\cline{2-5} & $R^2$ & \multicolumn{3}{c|}{0.96} \\
 \hline
\end{tabular}
\label{tab:parameter_linear_coeff_TSB}
\end{table}

We now perform a linear regression between the execution time of VUS, VUS$_{opt}$ and VUS$_{opt}^{mem}$, and $\alpha$, $\ell_a$ and $|T|$.
We report in Table~\ref{tab:parameter_linear_coeff_TSB} the slope coefficient for each parameter, as well as the $R^2$.  
The latter shows that the VUS$_{opt}$ and VUS$_{opt}^{mem}$ execution times are impacted by $\alpha$ at a larger degree than $\alpha$ affects VUS. 
On the other hand, the VUS$_{opt}$ and VUS$_{opt}^{mem}$ execution times are impacted to a significantly smaller degree by the time series length when compared to VUS. 
We also confirm that the anomaly length does not impact the execution time of VUS, VUS$_{opt}$, or VUS$_{opt}^{mem}$.
Finally, our experiments show that our optimized implementations VUS$_{opt}$ and VUS$_{opt}^{mem}$ significantly speedup the execution of the VUS measures (i.e., they can be computed within the same order of magnitude as R-AUC), rendering them practical in the real world.

\subsection{Summary of Results}

\begin{figure}[tb]
  \centering
  \includegraphics[height=7cm,width=\linewidth]{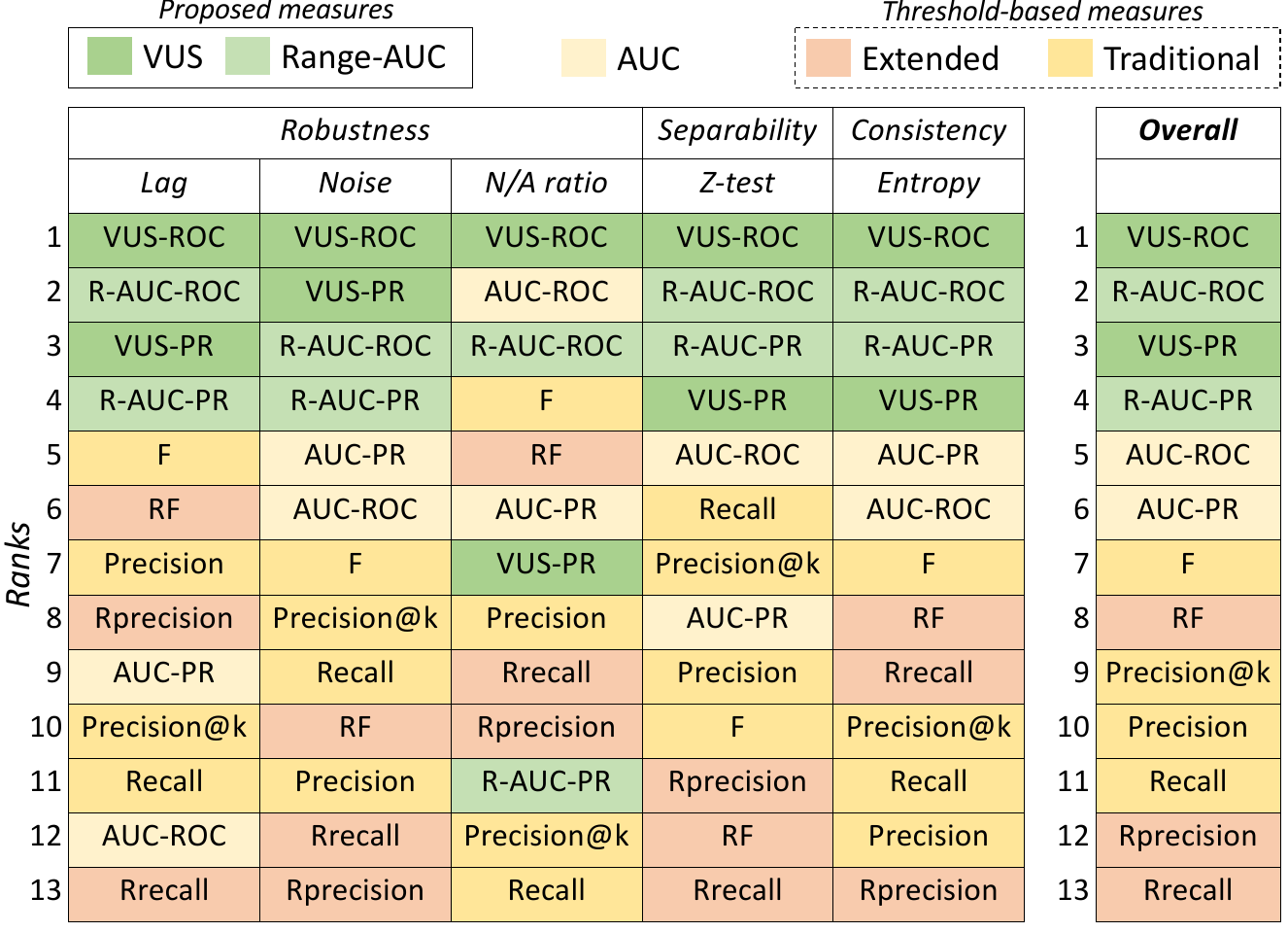}
  %\vspace*{-0.5cm}
  \caption{Ranks of the accuracy evaluation measures for the three different tests (i.e., on robustness, separability, and consistency) performed in the experimental evaluation, as well as the overall ranks (averaged for each measure on all tests). }
  \label{fig:overalltable}
\end{figure}

Figure~\ref{fig:overalltable} depicts the ranking of the accuracy measures for the different tests performed in this paper. The robustness test is divided into three sub-categories (i.e., lag, noise, and Normal vs. abnormal ratio). We also show the overall average ranking of all accuracy measures (most right column of Figure~\ref{fig:overalltable}).
Overall, we see that VUS-ROC is always the best, and VUS-PR and Range-AUC-based measures are, on average, second, third, and fourth. We thus conclude that VUS-ROC is the overall winner of our experimental analysis.

\commentRed{In addition, our experimental evaluation shows that the optimized version of VUS accelerates the computation by a factor of two. Nevertheless, VUS execution time is still significantly slower than AUC-based approaches. However, it is important to mention that the efficiency of accuracy measures is an orthogonal problem with anomaly detection. In real-time applications, we do not have ground truth labels, and we do not use any of those measures to evaluate accuracy. Measuring accuracy is an offline step to help the community assess methods and improve wrong practices. Thus, execution time should not be the main criterion for selecting an evaluation measure.}

\section{Conclusions}
\label{sec:conclusions}

Time-series AD is a challenging problem, and an active area of research. 
Given the multitude of solutions proposed in the literature, it is important to be able to properly evaluate them.
In this paper, we demonstrate the limitations of threshold-based accuracy measures. 
Even though AUC-based measures solve the threshold issues, we show that they cannot handle lag and noise. Overall, we show that the proposed VUS-based measures are more robust, and better separate accurate methods from inaccurate ones.

\commentRed{Despite the significant scalability improvement brought by $VUS_{opt}$ and $VUS_{opt}^{mem}$, the execution time is still higher than that of the simple AUC-based and the threshold-based approaches. 
Nevertheless, since the VUS-based measures are more robust, separable, and consistent, studying further optimization strategies is an important research direction.
Even though VUS-based methods are only relevant to the offline accuracy evaluation step, improving the execution time would benefit the relevant benchmarks.}

\begin{acknowledgements}
We thank the anonymous reviewers whose comments have greatly improved this manuscript. We also thank Yuhao Kang for his help during the early phase of this work. This research was supported in part by NetApp, Cisco Systems, Exelon Utilities, HPC resources from GENCI–IDRIS (Grants 2020-101471, 2021-101925), and EU Horizon projects AI4Europe (101070000), TwinODIS (101160009), ARMADA
(101168951), DataGEMS (101188416) and RECITALS (101168490).
\end{acknowledgements}

%\bibliographystyle{spphys}      % basic style, author-year citations
\bibliographystyle{spmpsci}      % mathematics and physical sciences
\bibliography{references}   % name your BibTeX data base

\end{document}